\documentclass{article}

 \usepackage[preprint]{neurips_2025}

\usepackage[utf8]{inputenc} 
\usepackage[T1]{fontenc}    
\usepackage{hyperref}       
\usepackage{url}            
\usepackage{booktabs}       
\usepackage{amsfonts}       
\usepackage{nicefrac}       
\usepackage{microtype}      
\usepackage{xcolor}         

\usepackage{adjustbox}
\usepackage{wrapfig}
\usepackage{bm} 
\usepackage{amsmath}
\usepackage{amssymb}
\usepackage{multirow}
\usepackage{colortbl}
\usepackage{makecell}
\definecolor{Gray}{gray}{.95}
\newcommand{\newshortname}{LLaVA-1.5c}

\usepackage{caption}  
\usepackage{algorithm}
\usepackage{algorithmic}

\title{LLaVA-c: Continual Improved Visual Instruction Tuning}

\author{Wenzhuo Liu$^{1,2}$, 
Fei Zhu$^{3}$, Haiyang Guo$^{1,2}$, Longhui Wei$^{4}$, Cheng-Lin Liu$^{1,2}$\thanks{Corresponding author.}~\\
$^1$School of Artificial Intelligence, UCAS\\
$^2$State Key Laboratory of Multimodal Artificial Intelligence Systems, CASIA\\
$^3$Centre for Artificial Intelligence and Robotics, HKISI-CAS \\
$^4$Huawei Inc \\
{\tt\small \{liuwenzhuo2020, zhufei2018, guohaiyang2023\}@ia.ac.cn,  liucl@nlpr.ia.ac.cn}
}

\begin{document}

\maketitle

\begin{abstract}
Multimodal models like LLaVA-1.5 achieve state-of-the-art visual understanding through visual instruction tuning on multitask datasets, enabling strong instruction-following and multimodal performance. However, multitask learning faces challenges such as task balancing, requiring careful adjustment of data proportions, and expansion costs, where new tasks risk catastrophic forgetting and need costly retraining. Continual learning provides a promising alternative to acquiring new knowledge incrementally while preserving existing capabilities. However, current methods prioritize task-specific performance, neglecting base model degradation from overfitting to specific instructions, which undermines general capabilities. In this work, we propose a simple but effective method with two modifications on LLaVA-1.5: spectral-aware consolidation for improved task balance and unsupervised inquiry regularization to prevent base model degradation.
We evaluate both general and task-specific performance across continual pretraining and fine-tuning. Experiments demonstrate that LLaVA-c consistently enhances standard benchmark performance and preserves general capabilities. \emph{For the first time, we show that task-by-task continual learning can achieve results that match or surpass multitask joint learning.} The code will be publicly released.
\end{abstract}

\section{Introduction}
\label{sec:intro}

\begin{wrapfigure}{r}{0.4\textwidth}
  \centering
  \vskip -0.3in
  \includegraphics[width=1.0\linewidth]{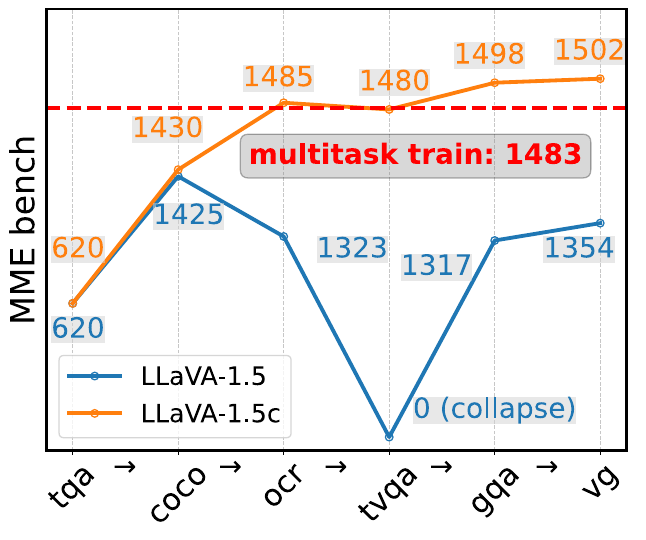}
  \vskip -0.13in
\caption{Results of training LLaVA-1.5 on LLaVA-665k~\cite{liu2024improved}. For the first time, we demonstrate that task-by-task continual learning can outperform joint learning on six tasks (e.g., coco, ocr).}
\vskip -0.4in
\label{fig:mme}
\end{wrapfigure}

Large Language Models (LLMs) like GPT \cite{brown2020language} and LLaMA \cite{touvron2023llama} excel in understanding instructions and generating accurate responses. 
To handle real-world multimodal data, which combines text, images, and other formats, the development of Multimodal Large Language Models (MLLMs) has drawn intensive attention.
LLaVA~\cite{llava} and LLaVA-1.5~\cite{liu2024improved} are state-of-the-art MLLMs that combine LLaMA with CLIP \cite{radford2021learning}, enabling visual understanding. Their success relies on visual instruction tuning, where the model is trained on a pre-collected, multitask visual instruction dataset covering tasks like visual question answering, visual grounding, and OCR (text recognition). This supervised fine-tuning paradigm equips the model with strong instruction-following capabilities and achieves impressive performance on multimodal tasks.
However, the \emph{multitask learning} paradigm (also known as joint training~\cite{kirkpatrick2017overcoming}) has two fundamental limitations when applied to real-world scenarios.
\textbf{(1) Task balancing}: MLLMs leverage data mixing strategy, in which balancing task-specific data is critical due to potential conflicts between datasets.  However, as task coverage expands, achieving the optimal configuration becomes exceedingly challenging.
\textbf{(2) Expansion costs}: In multitask training, learning new tasks independently can lead to catastrophic forgetting of existing knowledge. As a result, adding new capabilities needs retraining with both old and new data, which significantly increases time and storage costs.

Continual learning (CL) aims to enable models to acquire new knowledge while retaining existing knowledge. Unlike joint training, it dynamically adapts to new tasks, avoiding task-balancing issues inherent in multitask learning and enabling model capability expansion at a lower cost.
Despite its advantages, current CL methods face limitations: most approaches focus on visual classification tasks and often mitigate forgetting by restricting learning on new tasks (e.g., updating only a few parameters or learning prompts). In other words, these methods "learn less to forget less," significantly limiting their practical applicability.
\begin{figure*}[t]
  \centering
  \includegraphics[width=1.0\linewidth]{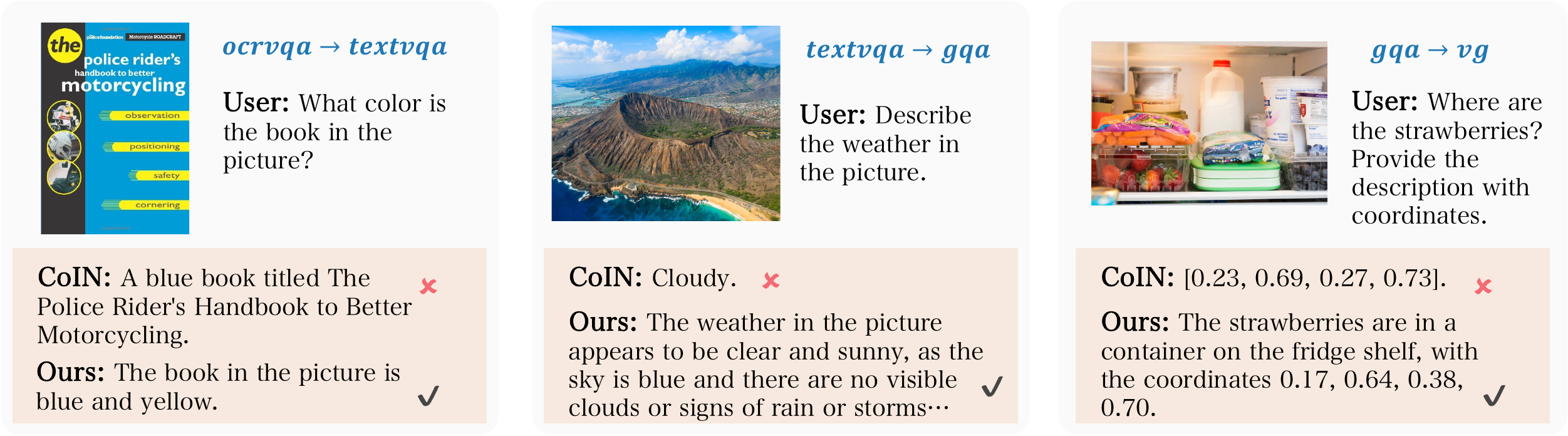}
  \vskip -0.05in
  \caption{Test examples on a recent MLLM-related continual learning method CoIN~\cite{chen2024coin}. It is found that exiting CL methods overfit task-specific instructions (Appendix \ref{app-details} and \ref{tab:datasets}). For instance, after incremental learning on GQA, model outputs single-word or phrase responses regardless of the input inquiry, losing the instruction-following ability. We term this phenomenon \textbf{\emph{Base Model Degradation}}, also observed in other methods~\cite{wang2023orthogonal,zeng2024modalprompt}.}
  \vskip -0.2in
  \label{fig:2}
\end{figure*}

Recently, a few MLLM-related CL methods have been proposed. However, as shown in Figure 2 and Table \ref{tab:datasets}, these models tend to overfit to task-specific instructions (e.g., "answer in a single word" or "generate a title") when learning new tasks, leading to a loss of their ability to handle diverse instructions and generate reliable responses. We term this phenomenon as \textbf{Base Model Degradation}. 
This exposes a key shortcoming in the evaluation of MLLM-related CL methods, which neglect the general performance while learning task-specific instructions.

In this work, we emphasize the importance of evaluating both the general capabilities of MLLMs (e.g., MME benchmarks) and task-specific performance (e.g., IconQA) in continual learning.
To address challenges related to task balancing in multitask training and the expansion costs of well-trained models, we consider two CL settings: continual pretraining and continual fine-tuning.
We propose LLaVA-c, a simple but effective continual learning method for MLLMs. Specifically, we introduce two improvements to LLaVA-1.5:
Firstly, we observe that model mixing is a good task balancer. Building on this insight, we propose Spectral-Aware Consolidation (SAC), which achieves improved task balance.
Secondly, we introduce Unsupervised Inquiry Regularization (UIR) into visual instruction tuning, effectively preventing base model degradation without hindering new‐task learning.
As shown in Figure \ref{fig:mme}, our method achieves remarkable results that even exceed the performance of multitask training.

Our main contributions are as follows:

\begin{itemize}
\item Continual learning offers a promising solution to achieve task balancing and alleviate expansion costs, but existing methods overlook base model degradation. This work is the first to evaluate both general and task-specific performance across continual pretraining and fine-tuning, respectively.
\item We propose a simple but effective continual learning approach named LLaVA-c with two key improvements to LLaVA-1.5: spectral-aware consolidation and unsupervised inquiry regularization.
\item Experiments show that LLaVA-c continually improves performance across benchmarks, prevents base model degradation, and achieves impressive results that match or surpass multitask learning.
\end{itemize}

\section{Relate Work}

\textbf{Multimodal Large Language Models.} Large language models (LLMs) have demonstrated exceptional capabilities in natural language understanding, generation, and in-context learning, enabling their application across diverse tasks~\cite{brown2020language, chowdhery2022palm, touvron2023llama}. 
Multimodal large language models (MLLMs) extend LLMs by integrating visual encoders and vision-language connectors to align image or video features with textual embeddings, enhancing multimodal instruction-following capabilities~\cite{radford2021learning, llava, li2023blip}. 
Representative works include LLaVA~\cite{llava}, MiniGPT-4~\cite{zhu2023minigpt4}, InstructBLIP~\cite{dai2023instructblip}, OpenFlamingo~\cite{alayrac2022flamingo}, and Emu~\cite{sun2023emu}. 
Among these, LLaVA and its variant LLaVA-1.5~\cite{liu2024improved} stand out for their simplicity in architecture. 
Despite variations in architecture and training strategies, most MLLMs rely on the joint training paradigm with multitask datasets. As tasks and capabilities grow, this paradigm faces inherent challenges in task balancing and capability expansion.
In this work, we build on LLaVA-1.5 and leverage continual learning to address these limitations, overcoming the constraints of the current multitask training paradigm in MLLMs.

\medskip

\textbf{Continual Learning.} Continual Learning (CL) aims to enable models to learn new tasks while retaining previously acquired knowledge. Existing CL strategies can be categorized into three main types: Regularization~\cite{li2017learning, kirkpatrick2017overcoming, Douillard2020SmallTaskIL}, prompt tuning~\cite{wang2022learning, smith2023coda}, and prefix tuning methods prevent forgetting by updating only a small subset of parameters or constraining parameter changes, but this limits the model’s ability to learn new tasks. Replay strategies~\cite{Rebuffi2017iCaRLIC, Wu2019LargeSI,gao2022r} mix old and new task data for joint training, which are simple and effective but face growing challenges with task conflicts and expansion costs as the number of tasks increases. Expansion strategies~\cite{yan2021dynamically,wang2022foster,hu2023dense} add new models at each stage to learn new tasks, which are impractical for large MLLMs due to excessive computational costs and model redundancy.

Recent MLLM continual learning methods~\cite{wang2023orthogonal,zeng2024modalprompt,chen2024coin} sequentially fine-tune a pre-trained LLaVA on downstream task datasets and evaluate task-specific performance. However, as shown in Figure 2, we observe a base model degradation phenomenon, where those models overfit to task-specific instructions after learning each new task. This overfitting deteriorates the capability of handling diverse instructions and producing reliable responses.
In this paper, we propose a simple and effective continual learning method based on LLaVA, as shown in Figure \ref{fig:2}. Our method retains general instruction-following capabilities while adapting to new tasks, preventing the base model degradation.

\section{Preliminary}
\subsection{Problem Formulation}
\textbf{Vision instruction tuning} optimizes MLLMs $f_{\bm{\theta}}: \mathcal{X} \to \mathcal{Y}$, mapping an input sequence $(x_1, x_2, \dots, x_n)$ to an output sequence $(y_1, y_2, \dots, y_m)$. 
The output space $\mathcal{Y} = \bigcup_{n=0}^N \mathcal{C}^n$ comprises text sequences up to length $N$ from vocabulary $\mathcal{C}$, while the input space $\mathcal{X} = \mathcal{V} \times \mathcal{Y}$ combines the image space $\mathcal{V}$ and $\mathcal{Y}$.
In the autoregressive framework, each output token $y_t$ satisfies $y_t = \arg\max_{c \in \mathcal{C}} p_{\bm{\theta}}\bigl(c \mid \bm{x}, y_1, y_2, \dots, y_{t-1}\bigr)$
, where $p_{\bm{\theta}}$ represents the conditional probability distribution.
The optimization objective is to minimize the $\mathcal{L}$ on the vision instruction dataset $\mathcal{D}$:
\begin{equation}
\label{eq2}
\begin{aligned}
\min_{\bm{\theta}} \mathcal{L}({\bm{\theta}}) = \mathbb{E}_{(\bm{x}, \bm{y}) \sim \mathcal{D}} \bigl[ \ell(f_{\bm{\theta}}(\bm{x}), \bm{y}) \bigr],
\end{aligned}
\end{equation}
where  $\ell(\cdot, \cdot)$ is the loss function (e.g., cross-entropy).

\textbf{Multitask learning} considers a dataset $\mathcal{D}$ composed of $T$ task-specific subsets ${\mathcal{D}_1, \mathcal{D}_2, \dots, \mathcal{D}_T}$. The Eq.~(\ref{eq2}) can be rewritten as:
\begin{equation}
\begin{aligned}
\min_{\bm{\theta}} \mathcal{L}({\bm{\theta}}) = \sum_{k=1}^T \lambda_k \cdot \mathbb{E}_{(\bm{x}, \bm{y}) \sim \mathcal{D}_t} \bigl[\ell(f_{\bm{\theta}}(\bm{x}), \bm{y}) \bigr],
\end{aligned}
\end{equation}
where $\lambda_k =\frac{\lvert \mathcal{D}_k \rvert}{\lvert \mathcal{D} \rvert}$ represents the proportion of subset $\mathcal{D}_k$.

\textbf{Continual learning} aims to minimize loss on the current task while preserving performance on past tasks. Formally, at task $t$, the optimization problem is:
\begin{equation} 
\begin{aligned} 
\min_{\bm{\theta}} \mathcal{L}({\bm{\theta}}) &= \mathbb{E}_{(\bm{x}, \bm{y}) \sim \mathcal{D}_k} \bigl[ \ell(f_{\bm{\theta}}(\bm{x}), \bm{y}) \bigr] + \sum_{i=1}^{t-1} \epsilon_i, \\
\text{s.t.}~~ \mathbb{E}_{(\bm{x},\bm{y}) \backsim \mathcal{D}_{i}}[ \ell(f_{\bm{\theta}}(\bm{x}), \bm{y}) &-\ell(f_{{\bm{\theta}}_{t-1}}(\bm{x}), \bm{y})] \leqslant \epsilon_{i},  \epsilon_{i} \geqslant 0; \forall i \in \{1,2,...,t-1\}.
\end{aligned} 
\end{equation} 
The term $\epsilon = {\epsilon_{i}}$ is a slack variable that allows a slight loss increase on previous tasks.

\subsection{MLLMs Continual Learning Setup}
LLaVA-1.5 training consists of two stages: vision-language alignment and visual instruction tuning. The first stage is unchanged in our experiment. The second stage uses the LLaVA-665k~\cite{liu2024improved} dataset.
As summarized in Table~\ref{tab:training_setup}, two continual learning settings are considered:

\begin{wraptable}[7]{r}{0.53\textwidth}
\centering
\vspace{-3ex}
\caption{Setup for LLaVA-1.5 under two CL settings.}
\vspace{-1ex}
\label{tab:training_setup}
\renewcommand{\arraystretch}{1.3}
\resizebox{0.53\textwidth}{!}{%
\begin{tabular}{p{2cm}|p{3.5cm}|p{4cm}}
\toprule
\textbf{Setting}             & \textbf{Initial Model}           & \textbf{Training Data Order}                                              \\ \midrule
Continual Pretraining       & Vicuna-7B, CLIP-L/14, Projector          & \textit{TQA} $\to$ \textit{COCO} $\to$ \textit{OcrQA} \newline $\to$ \textit{TextVQA} $\to$ \textit{GQA} $\to$ \textit{VG} \\ \midrule
Continual Fine-tuning      & LLaVA-1.5-7B                 & \textit{IconQA} $\to$ \textit{Super-clevr} \newline $\to$ \textit{Math} $\to$ \textit{Figure} $\to$ \textit{Arxiv}          \\ \bottomrule
\end{tabular}%
}
\vspace{-3ex}
\end{wraptable}

\textbf{Continual Pretraining:} 
The LLaVA-1.5 model, without visual instruction tuning, is sequentially trained on six domains from the 665k dataset: \textit{tqa}, \textit{coco}, \textit{ocr}, \textit{textvqa}, \textit{gqa}, and \textit{vg}. We evaluate the model on general benchmarks (e.g., MMbench, MME) to verify the continually improved general performance.

\textbf{Continual Fine-tuning:} The well-trained LLaVA-1.5 model is fine-tuned sequentially on datasets from five unseen domains. We evaluate the model on both general benchmarks (e.g., MME) and learned downstream tasks (e.g., IconQA).

\section{The Proposed Method: LLaVA-c}
\label{sec:method}
\begin{figure*}[t]
  \centering
  \includegraphics[width=0.95\linewidth]{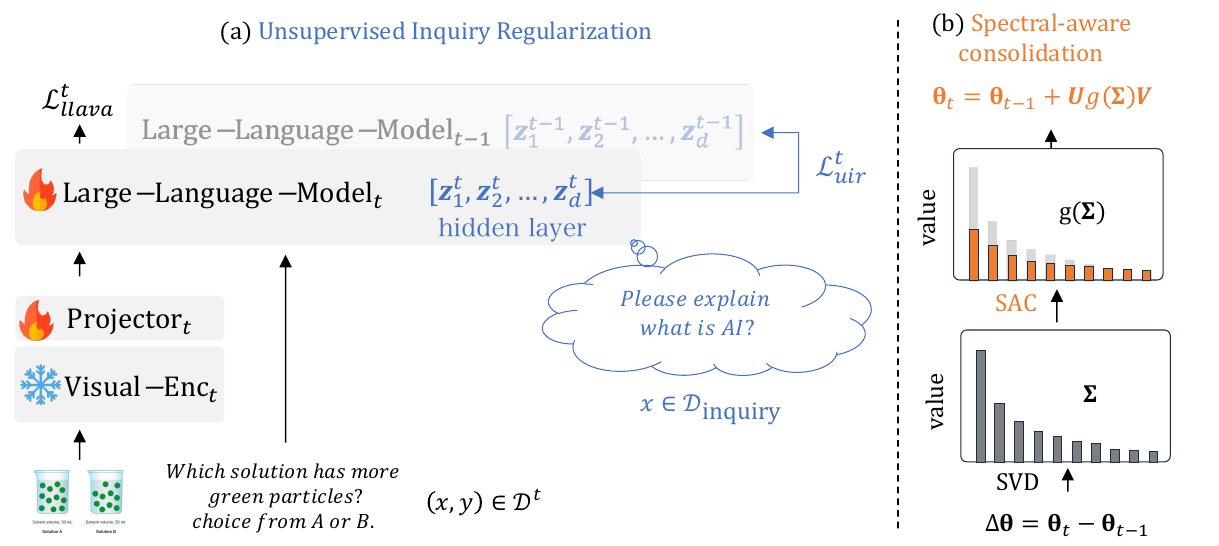}
  \vskip -0.05in
\caption{Illustration of LLaVA-c, which extends LLaVA-1.5 with two modifications: \textbf {Spectral-Aware Consolidation (SAC)} for better task balancing than ModelMix, and \textbf{Unsupervised Inquiry Regularization (UIR)} to prevent base model degradation and improve the performance of SAC. Our method is easy to implement and adds no extra hyperparameters during visual instruction tuning.}
  \vskip -0.2in
  \label{fig:3}
\end{figure*}

Multitask joint training is the dominant paradigm for MLLM training but faces two key challenges: task balancing and expansion cost.
To address these challenges, we explored continual learning to train MLLMs and extend their capabilities. 
As shown in Figure \ref{fig:3}, we introduced two modifications in LLaVA-1.5: \textbf{(1) Spectral-Aware Consolidation (SAC)}, which achieves improved task balance compared to model mixing; and \textbf{(2) Unsupervised Inquiry Regularization (UIR)}, which stabilizes the language feature space, effectively preventing base model degradation and further enhancing the performance of SAC.
We call the improved model LLaVA-c.

\subsection{Spectral-Aware Consolidation}
A primary problem in the continual learning of MLLMs is mitigating the forgetting of previously learned knowledge, including general and task-specific knowledge. To this end, we explore model mixing (ModelMix) to consolidate the knowledge as follows: $\bm{\widetilde{\theta}}_{t} = \alpha \bm{\theta}_{t} + (1 - \alpha) \bm{\widetilde{\theta}}_{t-1}$, where $\bm{\widetilde{\theta}}_{t-1}$ and $\bm{\widetilde{\theta}}_{t}$ denote the averaged model at the end of stage $t-1$ and $t$, respectively. The hyper-parameter $\alpha \in [0,1]$ determines the weighting of the old and new models.

\begin{figure}[t]
\vskip -0.2in
\centering
\includegraphics[width=1.0\linewidth]{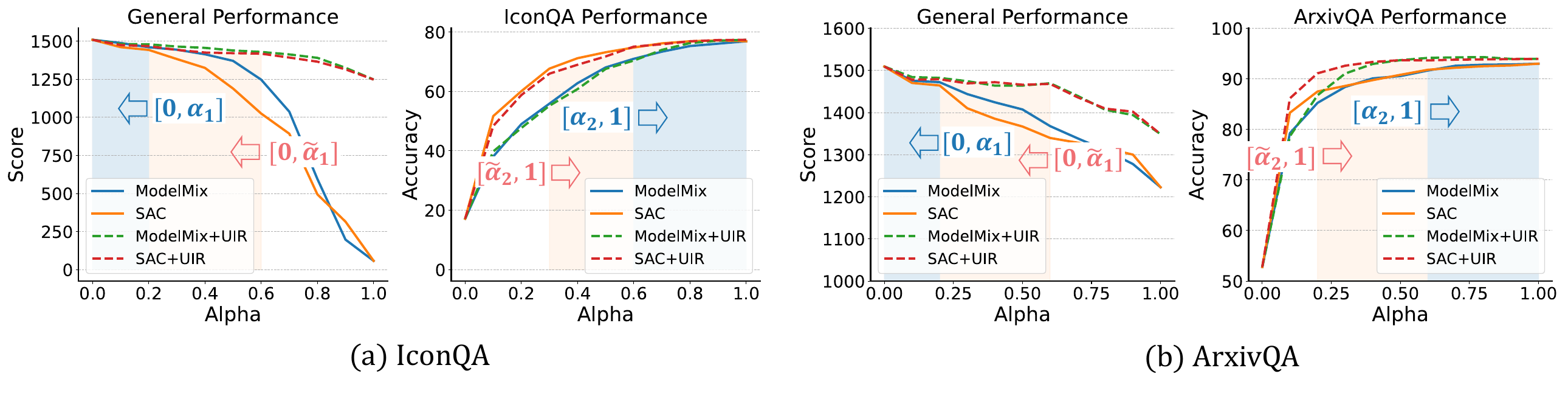}
\vskip -0.1in
\caption{Impact of ModelMix ratio $\alpha$ in fine-tuning LLaVA-1.5 on IconQA and ArxivQA. SAC improves task balance within $[0, \alpha_1]$, while combining UIR and SAC yields the best performance, forming an Ideal Region $[\tilde{\alpha}_2,\tilde{\alpha}_1]$ for continual learning.}
\label{fig:modelmix}
\vskip -0.2in
\end{figure}

\textbf{Model mixing is a good task balancer.} 
As shown in Figure \ref{fig:modelmix}, ModelMix exhibits a symmetric phenomenon, revealing a potential win-win effect during continual learning.
\begin{itemize}
    \item  General-stable Region: As $\alpha$ increases in [0, $\alpha_1$], ModelMix enhances task-specific performance and preserves general capabilities.
    \item Task-stable Region:  As $\alpha$ decreases in [$\alpha_2$, 1], it improves general performance while maintaining task-specific knowledge.
\end{itemize}
Specifically,  if $\alpha_2$ is smaller than $\alpha_1$, new knowledge can be integrated without forgetting existing capabilities, i.e. [$\alpha_2$, $\alpha_1$] is the \textbf{Ideal Region} for continual learning.

However, direct model mixing rarely forms an Ideal Region, as $\alpha_2$ typically exceeds $\alpha_1$. This naturally raises the question: can we expand both the General and Task regions?
To this end, we delve into the mechanism of ModelMix by providing a novel spectral view. 
Specifically, it adjusts the parameter update $\Delta \bm{\theta}$ as follows: 
\begin{equation} 
\bm{\widetilde{\theta}}_t = \bm{\widetilde{\theta}}_{t-1} + \alpha (\bm{{\theta}}_t - \bm{\widetilde{\theta}}_{t-1}) = \bm{\widetilde{\theta}}_{t-1} + \alpha \Delta \bm{\theta}.
\end{equation} 
Applying singular value decomposition (SVD) \footnote{Before performing SVD decomposition, we reshape the parameters to adjust their shape so that $\Delta \bm{\theta} \in \mathbb{R}^{M \times N}$.}, ModelMix can be viewed as scaling the singular values of $\Delta \bm{\theta}$:
\begin{equation} 
\alpha \Delta \bm{\theta} = \mathbf{U} (\alpha \mathbf{\Sigma}) \mathbf{V}^\top = \mathbf{U} g(\mathbf{\Sigma}) \mathbf{V}^\top,
\end{equation}
where $ \mathbf{\Sigma} = \text{diag}(\sigma_1, \sigma_2, \dots, \sigma_r)$ with $\sigma_1 \ge \sigma_2,..., \ge \sigma_r$. 

From the spectral perspective, ModelMix applies a uniform scaling function $g(\cdot)=\alpha$ across eigenvector directions in low-dimensional space of $\Delta \bm{\theta}$. Intuitively, their impact on old parameters is negligible for smaller singular values, so reducing their scaling helps retain task-specific information without forgetting.

Based on this intuition, we propose a spectral-aware consolidation (SAC) function $g(\cdot)$ that accounts for subspace magnitude in the consolidation process, implemented via sliding window average \footnote{When the window extends beyond the rank limit (i.e., $i + k > r$), we handle boundary cases by padding with the last singular value $\sigma_{r}$.}.
\begin{equation}
\begin{aligned} 
 g(\mathbf{\Sigma}) &= \text{diag}\bigl(\alpha_1 \sigma_1, \alpha_2 \sigma_2, \dots, \alpha_r \sigma_r\bigr)= \frac{1}{k}\text{diag}\bigl(\Sigma_{i=1}^{k+1}\sigma_{i}, \Sigma_{i=2}^{k+2}\sigma_{i}, \dots, \Sigma_{i=r}^{k+r}\sigma_{i}),
\end{aligned}
\end{equation}
where $\alpha_i$ ($i=1,..,r$) is adjusted dynamically based on $\sigma_i$, e.g., larger $\sigma_i$ values receive smaller $\alpha_i$. The sliding window size $k$ is computed using the following optimization objective, adding no extra hyperparameters compared to vanilla ModelMix.
\begin{equation}
\begin{aligned} 
 \min_{k}(\frac{1}{k}\text{diag}\Sigma_{i=1}^{k+1}\sigma_{i}-\alpha\sigma_{1})^2,\text{s.t.}~~ \forall k &\in \{1,2,...,r\}.
\end{aligned}
\end{equation}

\begin{figure}[t]
  \centering
  \vskip 0.05in
  \begin{minipage}[t]{0.49\linewidth}
    \centering
    \includegraphics[width=\linewidth]{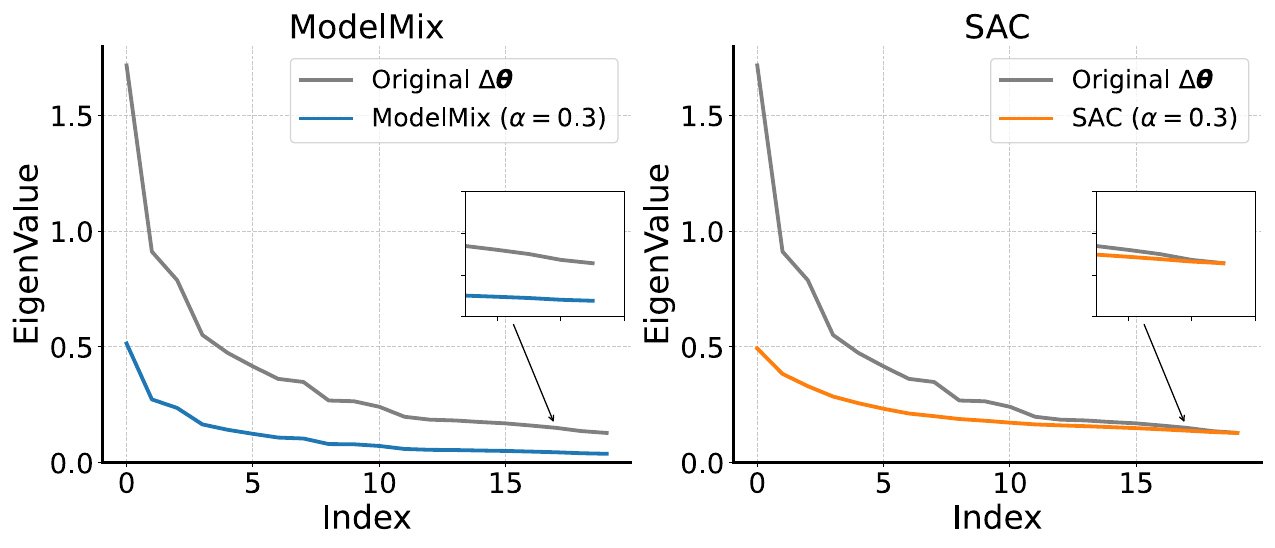}
    \vskip -0.05in
    \caption{Spectral comparison of ModelMix and SAC. ModelMix applies uniform scaling to all eigenvectors, whereas SAC adaptively amplifies directions with larger eigenvalues.}
    \label{fig:sac}
  \end{minipage}%
  \hfill
  \begin{minipage}[t]{0.49\linewidth}
    \centering
    \includegraphics[width=\linewidth]{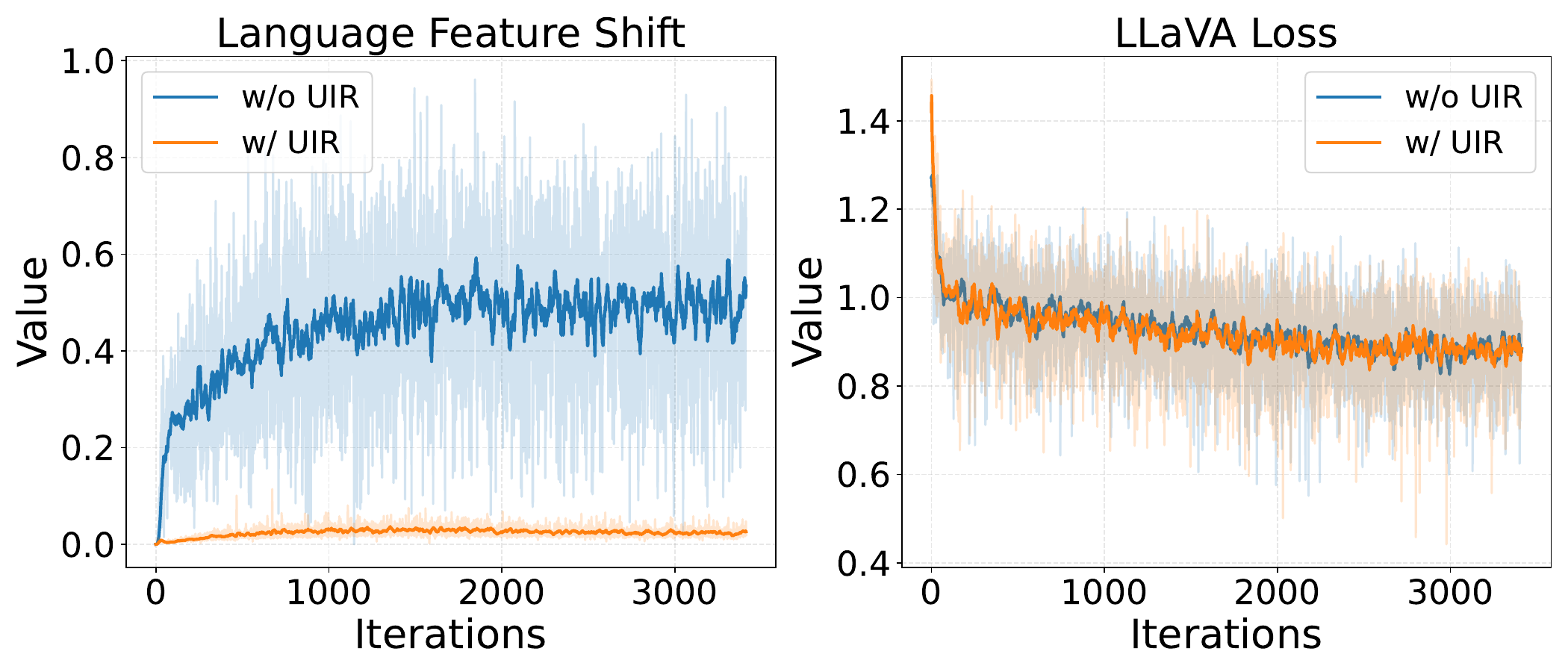}
    \vskip -0.05in
    \caption{Loss curve comparison in task-specific learning. Applying UIR to unlabeled inquiries mitigates language shift while keeping the LLaVA loss aligned with the vanilla trend.}
    \label{fig:loss_curves}
  \end{minipage}
  \vskip -0.2in
\end{figure}

In Figure \ref{fig:sac}, we compare the eigenvalue curves of ModelMix and SAC under the same mix ratio ($\alpha=0.3$). As expected, the results show that the sliding window average effectively enables adaptive scaling based on eigenvalue magnitude.
Furthermore, as shown in Figure \ref{fig:modelmix}, experiments on IconQA and ArxivQA validate the practical effectiveness of SAC. 
It achieves a better balance in the General-Stable Range $[0,\alpha_1]$, remarkably enhancing task-specific performance with negligible impact on general performance.

\subsection{Unsupervised Inquiry Regularization}
Section 4.1 demonstrates that SAC enhances task-specific performance within the General-Stable Region, thereby expanding $\left[\alpha_2, 1\right]$.
We argue that preventing Base Model Degradation is key to maintaining general capabilities, which is crucial for expanding the $\left[0, \alpha_1\right]$ region.

For instance, as shown in Figure \ref{fig:2}, after learning an image-caption task, the model overfits generating captions even for unrelated inquiries, leading to degraded performance on tasks requiring natural responses and general knowledge.
To delve into this problem, we provide a quantitative analysis by tracking the dynamics of text representations in the deep feature space. Specifically, given some unlabeled, complex text inquiry, we compute the language shift defined by $L_2$ distance in the deep feature space. As shown in Figure \ref{fig:loss_curves}, a significant shift can be observed during the continual learning process. Therefore, we propose a simple yet effective unsupervised inquiry regularization (UIR) to:
\begin{itemize} 
\item Prevents overfitting to task-specific instructions to avoid the base model degradation. 
\item Stabilize the text-related feature space to reduce language shifts, leading to homogenized model parameters in different continual learning stages.
\end{itemize}
As illustrated in Figure \ref{fig:3}, we introduce a small surgery dataset containing easily available and unlabeled inquiry text, and stabilize the representations as follows:
\begin{equation}
\label{eq-uir}
\begin{aligned}
\mathcal{L}^{t}_{\text{uir}} = \mathbb{E}_{\bm{x} \sim \mathcal{D}_{\text{inquiry}}} \left\| h\bigl[f_{\bm{\theta}_{t}}(\bm{x}) \bigr] - h\bigl[f_{\bm{\theta}_{t-1}}(\bm{x}) \bigr] \right\|_2,
\end{aligned}
\end{equation}
where $h[f_{\theta}(\bm{x})]$ is the output of the final hidden layer, $|| \cdot ||_2$ denotes the $L_2$-norm, and $\mathcal{D}_{\text{inquiry}}$ is a surgery instruction dataset with unlabeled inquiry samples.

We highlight that UIR only applies to unlabeled text data, rather than the multimodal training data. In this way, learning new tasks is not affected, making our method remarkably different from existing continual learning methods that reduce forgetting mainly by learning less new knowledge. 
The results in Figure \ref{fig:loss_curves} show that the LLaVA loss curve of our approach closely follows the vanilla trend, which differs from previous methods. As a result, general knowledge performance can be significantly improved, and the new task performance can be maintained or even enhanced during knowledge consolidation, as shown in Figure \ref{fig:modelmix}. 

\subsection{Final Learning Objective}
The original LLaVA loss ~\cite{liu2024improved} is used to learn new task knowledge:
$
\mathcal{L}^t_{\text{llava}} = \mathbb{E}_{(\bm{x}, \bm{y}) \sim \mathcal{D}_t} \bigl[ \ell(f_{\bm{\theta}_t}(\bm{x}), \bm{y}) \bigr].
$
Therefore, in the $t$-th continual learning stage, the total training objective is:
\begin{equation}
\label{eq-llava-all}
\mathcal{L}^t = \mathcal{L}^t_{\text{llava}} + \mathcal{L}^t_{\text{uir}}.
\end{equation}
Compared to the original LLaVA/LLaVA-1.5 models, the proposed UIR does not introduce any extra hyperparameters, making it easy to implement and apply without complex tuning. After each continual stage, spectral-aware consolidation integrates the newly learned knowledge. The total process is summarized in  Appendix \ref{app-code} and Algorithm \ref{alg:llava-c}.

\section{Experiments}
\label{sec:exp}
\noindent\textbf{Benchmarks and comparison methods.} We assessed the performance of LLaVA-1.5c under two settings: continual pretraining and fine-tuning. The evaluation included general benchmarks like POPE~\cite{li2023pope}, MME~\cite{fu2023mme}, MMBench~\cite{liu2023mmbench}, SEED-Bench~\cite{li2023seed}, LLaVA-Bench~\cite{llava} , MM-Vet~\cite{yu2023mmvet}, and MMMU \cite{yue2024mmmu}, as well as task-specific benchmarks such as Super-clevr~\cite{li2023super}, IconQA~\cite{lu2021iconqa}, Clevr-Math~\cite{dahlgren2022clevr}, ArxivQA~\cite{li-etal-2024-multimodal-arxiv}, and FigureQA~\cite{kahou2017figureqa}. 
The comparison methods include: (1) traditional CL methods such as LWF~\cite{li2017learning}, EWC~\cite{kirkpatrick2017overcoming}, and L2P~\cite{wang2022learning}; (2) recent MLLM/LLM-related CL methods like O-LoRA~\cite{wang2023orthogonal}, Modal-prompt~\cite{zeng2024modalprompt}, and CoIN~\cite{chen2024coin}.

\medskip
\noindent\textbf{Implementation details.} LLaVA-1.5c is run on PyTorch using 8 NVIDIA L20 GPUs with 8 batches per GPU, strictly following LLaVA-1.5~\cite{liu2024improved} settings without additional hyperparameters.
Specifically, LLaVA-1.5 employs Vicuna-7B/13B~\cite{zheng2023judging} language models and CLIP-L/14@336 vision models~\cite{radford2021learning}. 
The training consists of two phases: the first with a learning rate of 2e-3 and the second at 2e-5. We use 2x gradient accumulation to maintain a batch size of 128.
After training, LLaVA-c employs SAC to integrate knowledge, setting alpha at 0.2 by default. Further implementation details can be found in the Appendix \ref{other-details}.

\begin{table*}[t!]
\vskip -0.1in
\centering
\setlength\tabcolsep{4.5pt}
\renewcommand{\arraystretch}{1.05}
\caption{\textbf{Comparison of Continual Pretraining}: We compare (1) SoTA MLLM methods with multitask pretraining and (2) LLaVA-1.5 models using different continual learning methods. $^\dagger$ denotes reproduced results, and other MLLM results are from LLaVA-1.5~\cite{liu2024improved}. Our method achieves continual pretraining performance that matches or surpasses multitask pretraining, whereas previous continual learning methods perform poorly in this setting.}
\vskip -0.03in
\resizebox{0.97\textwidth}{!}{
\begin{tabular}{l l| c c  c | c| c  c| c c c | c | c }
\toprule
\multirow{2}{*}{Method} & \multirow{2}{*}{LLM} & \multicolumn{3}{c|}{POPE} & MME & \multicolumn{2}{c|}{MMBench} & \multicolumn{3}{c|}{SEED-Bench} & LLaVA- & MM-Vet \\
 &  & rand & pop & adv &  & en & cn & all & img & vid & Wild &  \\
\midrule
\multicolumn{13}{l}{\textbf{\textit{multitask pre-train:}}} \\
\midrule
BLIP2-14B & Vicuna-13B & \textbf{89.6} & \underline{85.5} & \underline{80.9} & 1293.8 & -- & -- & 46.4 & 49.7 & 36.7 & 38.1 & 22.4 \\
InstructBLIP-8B & Vicuna-7B & -- & -- & -- & -- & 36 & 23.7 & 53.4 & 58.8 & 38.1 & \underline{60.9} & \underline{26.2} \\
InstructBLIP-14B & Vicuna-13B & \underline{87.7} & 77 & 72 & 1212.8 & -- & -- & -- & -- & -- & 58.2 & 25.6 \\
Shikra-13B & Vicuna-13B & -- & -- & -- & -- & 58.8 & -- & -- & -- & -- & -- & -- \\
IDEFICS-9B & LLaMA-7B & -- & -- & -- & -- & 48.2 & 25.2 & -- & 44.5 & -- & -- & -- \\
IDEFICS-80B & LLaMA-65B & -- & -- & -- & -- & 54.5 & 38.1 & -- & 53.2 & -- & -- & -- \\
Qwen-VL & Qwen-7B & -- & -- & -- & -- & 38.2 & 7.4 & 56.3 & 62.3 & \underline{39.1} & -- & -- \\
Qwen-VL-Chat & Qwen-7B & -- & -- & -- & \textbf{1487.5} & \underline{60.6} & \underline{56.7} & \underline{58.2} & \underline{65.4} & 37.8 & -- & -- \\
LLaVA-1.5$^\dagger$ & Vicuna-7B & 87.4 & \textbf{86.2} & \textbf{84.5} & \underline{1483.5} & \textbf{65.9} & \textbf{58.6} & \textbf{61.5} & \textbf{67.3} & \textbf{39.3} & \textbf{66.4} & \textbf{32.6} \\
\midrule
\multicolumn{13}{l}{\textbf{\textit{continual pre-train:}}} \\
\midrule
\rowcolor{Gray}
\newshortname{} & Vicuna-7B & \textbf{87.7} & \textbf{86.2} & \textbf{85.1} & \textbf{1502.4} & \textbf{65.2} & \textbf{58.8} & \textbf{61.0} & \textbf{66.4} & \textbf{40.7} & \textbf{66.8} & \textbf{31.2} \\
Finetune & Vicuna-7B & 83.8 & 82.8 & 81.6 & 1354.9 & 49.5 & 44.1 & 35.0 & 37.7 & 24.8 & 24.8 & 25.0 \\
LWF & Vicuna-7B & 83.3 & 82.5 & 81.0 & 1342.3 & 48.8 & 43.7 & 34.5 & 37.2 & 24.0 & 35.6 & 24.4 \\
EWC & Vicuna-7B & 82.1 & 81.1 & 80.4 & 1320.8 & 46.1 & 43.4 & 33.3 & 35.5 & 24.9 & 39.2 & 24.1 \\
L2P & Vicuna-7B & 78.2 & 74.3 & 73.5 & 1090.7 & 43.5 & 34.8 & 28.1 & 28.6 & 26.3 & \underline{45.1} & 23.5 \\
O-LoRA & Vicuna-7B & \underline{85.5} & \underline{83.2} & \underline{82.7} & \underline{1380.3} & \underline{53.8} & \underline{49.4} & 39.5 & 41.7 & \underline{30.7} & 36.7 & \underline{28.4} \\
CoIN & Vicuna-7B & 84.5 & 83.0 & 81.8 & 1367.4 & 51.8 & 45.7 & \underline{40.7} & \underline{43.8} & 28.8 & 29.4 & 26.8 \\
\bottomrule
\end{tabular}
}
\vskip -0.1in
\label{tab:pretrain}
\end{table*}
\subsection{Main Results}
\label{sec:4.1}

\begin{table}[t]
\vskip -0.05in
\centering
\setlength\tabcolsep{4.5pt}
\renewcommand{\arraystretch}{1.05}
\caption{Evaluation on MMMU benchmark of continual pretraining}
\vskip 0.03in
\label{app:mmmu}
\resizebox{0.85\linewidth}{!}{%
  \begin{tabular}{l c c c c c c c}
    \toprule
    \multirow{2}{*}{Method}
      & \multicolumn{6}{c}{Discipline Accuracy (\%)} 
      & \multirow{2}{*}{Overall} \\
    \cmidrule(lr){2-7}
      & Art \& Design
      & Business
      & Science
      & \shortstack{Health \\ \& Medicine}
      & \shortstack{Humanities \\ \& Social Sci.}
      & \shortstack{Tech \\ \& Engineering}
      &  \\
    \midrule
    Multi-task    & 50.8 & 26.0 & 26.7 & 33.3 & 47.5 & 30.0 & 34.4 \\
    Fine-tune     & 34.2 & 17.3 & 15.0 & 18.2 & 31.7 & 21.1 & 22.1 \\
    O-LoRA        & \underline{40.8} & \underline{21.7} & 20.0 & \underline{25.3} & \underline{35.8} & \underline{25.9} & \underline{27.4} \\
    CoiN          & 38.5 & 20.3 & \underline{22.5} & 23.3 & 34.2 & 23.4 & 26.2 \\
    \rowcolor{Gray}
    LLaVA-1.5c           & \textbf{54.2} & \textbf{27.3} & \textbf{29.3} & \textbf{32.0} & \textbf{50.8} & \textbf{30.0} & \textbf{35.8} \\
    \bottomrule
  \end{tabular}%
}
\vskip -0.15in
\end{table}

\noindent\textbf{Continual Pretraining.} 
As shown in Table \ref{tab:pretrain} and \ref{app:mmmu}, LLaVA-1.5 exhibits a significant performance gap between continual pretraining and multitask pretraining, with up to a 2.7x difference on LLaVA-Bench.
However, traditional continual learning methods perform poorly in this setting because they mitigate forgetting by sacrificing performance on new tasks.
In particular, methods like L2P, which keep the model fixed and only learn prompts, perform worse than fine-tuning LLaVA models directly.
\begin{wrapfigure}[31]{r}{0.55\textwidth}  
  \centering
  \vspace{-0.1in}
  \begin{minipage}{\linewidth}
    \centering
    \captionsetup{type=table}  
    \caption{\textbf{Comparison of Continual Fine-tuning}: We perform continual fine-tuning on the well-trained LLaVA-1.5-7B model following the task order \textit{Icon} → \textit{Super} → \textit{Math} → \textit{Fig} → \textit{Arxiv}, and evaluate the final model on all learned tasks.}
    \label{tab:finetune}
    \setlength\tabcolsep{6pt}
    \renewcommand{\arraystretch}{1.05}
    \resizebox{0.95\linewidth}{!}{
      \begin{tabular}{l | c c c c c c }
        \toprule
        \textbf{Method} & \rotatebox{60}{\textbf{IconQA}} & \rotatebox{60}{\textbf{Super}} & \rotatebox{60}{\textbf{Math}} & \rotatebox{60}{\textbf{Figure}} & \rotatebox{60}{\textbf{Arxiv}} & \rotatebox{60}{\textbf{Avg}} \\
        \midrule
        Zero-shot   & 17.50 & 17.85 & 19.25 & 17.0  & 52.80 & 24.9 \\
        \midrule
        \rowcolor{Gray}
        LLaVA-1.5c  & \textbf{65.1}  & \textbf{46.0}  & \textbf{45.9}  & 44.1  & 90.4  & \textbf{58.3} \\
        Finetune    & 11.3  & 35.1  & 27.5  & 38.5  & 92.6  & 41.0 \\
        LWF         & 45.3  & 37.1  & 31.4  & 41.8  & 91.0  & 49.3 \\
        EWC         & 48.1  & 34.5  & 32.5  & 44.7  & 89.3  & 49.8 \\
        L2P         & 52.0  & 39.9  & 41.0  & 39.5  & 86.3  & 51.7 \\
        M-Prompt    & \underline{55.9}  & 33.5  & \underline{43.9}  & 42.3  & 87.7  & 52.6 \\
        O-LoRA      & 38.6  & \underline{44.8}  & 43.0  & \textbf{49.0}  & \textbf{94.5}  & \underline{54.0} \\
        CoIN        & 10.4  & 41.6  & 36.5  & \underline{45.9}  & \underline{94.0}  & 45.7 \\
        \bottomrule
      \end{tabular}
    }
    \vspace{1ex}

    \captionsetup{type=figure}  
    \includegraphics[width=\linewidth]{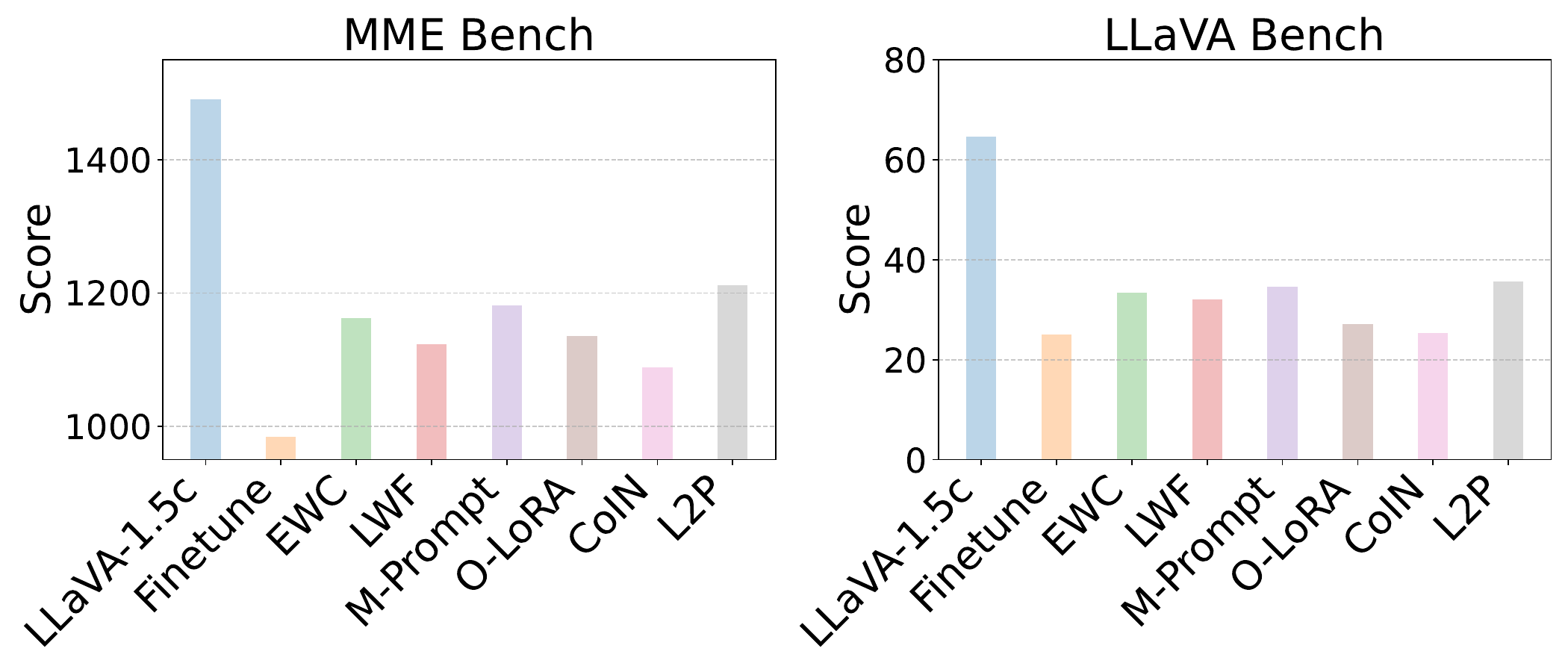}
    \caption{Comparison of general performance after continual fine-tuning. Previous methods reduce forgetting across downstream tasks but suffer significant degradation in general performance.}
    \label{fig:ft-genernal}
  \end{minipage}
\end{wrapfigure}
Although recent MLLM-based methods show improvements, they still exhibit a significant gap compared to multitask pretraining. Furthermore, our qualitative analysis of methods like CoIN reveals that they fail to address task-specific instruction overfitting.
In contrast, LLaVA-c significantly outperforms previous methods. 
For instance, on benchmarks like MME and POPE, our method even surpasses multitask learning and achieves comparable performance to others, such as MM-Vet.
Additionally, qualitative results in Figure \ref{fig:2} demonstrate clear improvements, as our method preserves instruction-following capabilities and prevents base model degradation effectively.

\noindent\textbf{Continual Fine-tuning.} 
In this setting, we employ continual learning to extend a well-trained MLLM to unseen tasks at a low cost. As shown in Table \ref{tab:finetune}, the LLaVA-1.5 model (pre-trained on 665k multitask data) performs poorly on some unseen tasks, such as IconQA, achieving a zero-shot accuracy of only 17.5\%.
Previous methods outperform direct fine-tuning under the continual fine-tuning setup, with O-LoRA, L2P, and M-Prompt achieving the best results. However, these methods lose their original instruction-following capabilities when adapting to new tasks, as demonstrated by the general performance evaluations in Figure \ref{fig:ft-genernal}. For example, while O-LoRA achieves high scores on downstream tasks, its performance on LLaVA Bench decreases by 60\%.
In contrast, our method achieves significant improvements in both general and task-specific performance. After fine-tuning on five datasets, LLaVA Bench performance remains at 97\% of its original level.

\subsection{Ablation Study and Further Analysis}
\textbf{Impact of components in LLaVA-c.}
As shown in Figure \ref{fig:modelmix}, we conducted 1-step incremental learning ablation experiments on IconQA and ArxivQA. UIR significantly mitigates base model degradation without sacrificing new task performance, which is distinct from previous methods.
\begin{wraptable}[11]{r}{0.5\textwidth}
\vspace{-2ex}
\caption{Impact of each component in LLaVA-c. We perform 3-step continual fine-tuning following the task order \textit{Icon} $\to$ \textit{Super} $\to$ \textit{Arxiv} and evaluate final model on general and task performance.}
\centering
\setlength\tabcolsep{6pt}
\renewcommand{\arraystretch}{1.05}
\resizebox{0.5\textwidth}{!}{
\begin{tabular}{l | c c | c c c }
\toprule
\multirow{2}{*}{\textbf{Method}} & \multicolumn{2}{c|}{\textbf{General}} & \multicolumn{3}{c}{\textbf{Task-specific}} \\
 & L-bench & MME & Icon & Super & Arxiv \\
\midrule
Vanilla & 28.1 & 712.2 & 12.6 & 38.2 & \underline{92.8} \\
+ SAC & 46.5  & \underline{1329.5} & \underline{52.1} & 40.4 & 85.3 \\
+ UIR & \underline{55.3} & 1276.4 & 44.5 & \underline{45.2} & \textbf{94.2} \\
\rowcolor{Gray}
+ SAC \& UIR & \textbf{66.8} & \textbf{1461.4} & \textbf{65.8} & \textbf{49.1} & 91.9 \\
\bottomrule
\end{tabular}
}
\label{tab:compnent}
\vskip -0.1in
\end{wraptable}
Moreover, on ArxivQA, UIR not only prevents degradation but also improves new task performance, outperforming direct fine-tuning.
Additionally, SAC achieves better task balance. Within the range of 0 to $\alpha_1$, SAC substantially enhances new task performance while maintaining comparable general performance.
We also performed 3-step incremental learning ablation experiments, as shown in Table \ref{tab:compnent}. These results demonstrate that the combination of UIR and SAC yields the best overall performance.

\begin{figure}[t]
  \centering
  \begin{minipage}[t]{0.49\textwidth}
    \centering
    \includegraphics[width=\linewidth]{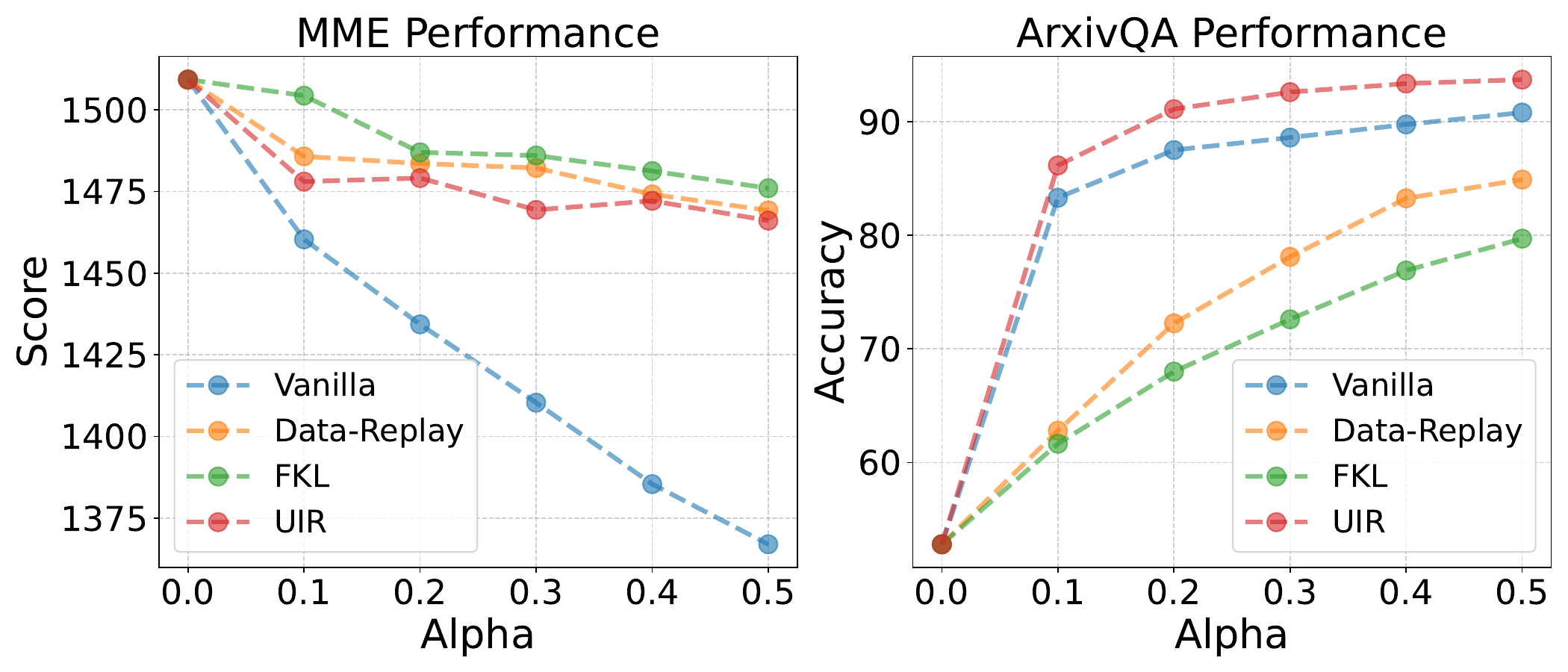}
    \vskip -0.05in
    \caption{Comparison of data replay and forward KL loss in a 1-step incremental learning experiment on ArxivQA.}
    \label{fig:data-replay}
  \end{minipage}%
  \hfill
  \begin{minipage}[t]{0.49\textwidth}
    \centering
    \includegraphics[width=\linewidth]{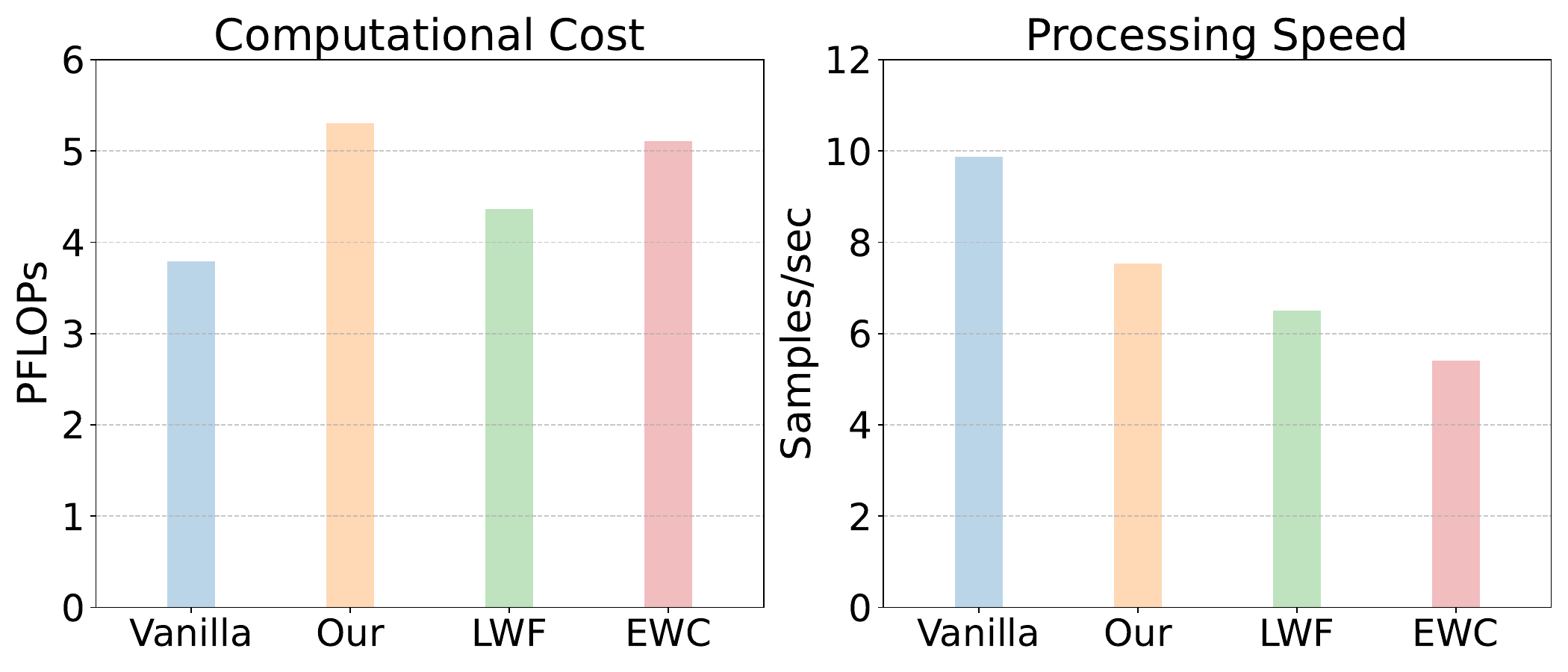}
    \vskip -0.05in
    \caption{Comparison of training FLOPs and time. Our method is comparable to previous continual learning approaches.}
    \label{fig:flops}
  \end{minipage}
    \vskip -0.2in
\end{figure}

\medskip
\textbf{Comparison with data-replay and FKL.}
We further ablate UIR by comparing two strategies: (1) Data-replay: Adding supervised information to the unsupervised instruction set in UIR and jointly training it with ArxivQA. (2) Forward Kullback-Leibler divergence (FKL): which replaces MSE loss in UIR with FKL loss and aligns new and old model logits.
As shown in Figure \ref{fig:data-replay}, we perform a 1-step fine-tuning on ArxivQA using LLaVA-1.5-7B and integrate knowledge via SAC.
Data-replay outperforms FKL by incorporating additional ground truth labels. 
However, both methods reduce general knowledge forgetting at the cost of new task learning. In contrast, UIR effectively reduces general knowledge forgetting while maintaining or even improving new task learning, demonstrating a clear advantage.

\medskip
\textbf{Evaluation on incremental stage.}
In Table \ref{tab:task_order_performance}, we evaluate the model at each incremental stage of continual fine-tuning. The results show that our method maintains strong general performance and effectively preserves task-specific performance. For some tasks,
\begin{wraptable}[10]{r}{0.5\textwidth}
\vspace{-2ex}
\caption{Task-specific and general performance at each continual fine-tuning stage.}
\vspace{-1ex}
\centering
\setlength\tabcolsep{6pt} 
\renewcommand{\arraystretch}{1.1} 
\resizebox{0.5\textwidth}{!}{
\begin{tabular}{l | c c c c c | c}
\toprule
\multirow{2}{*}{\textbf{Order}} & \multicolumn{5}{c|}{\textbf{Task-specific Performance}} & \multirow{2}{*}{\textbf{MME}} \\
& Icon & Super & Math & Fig & Arxiv & \\
\midrule
Zero-shot & 17.5 & 17.9 & 19.3 & 17.0 & 52.8 & \textbf{1509.3} \\
IconQA    & 66.0 & 32.3 & 21.0 & 17.9 & 56.8 & 1443.0 \\
Super     & \textbf{66.3} & \textbf{49.2} & 20.6 & 18.1 & 57.6 & 1451.0 \\
Math      & \textbf{66.3} & \underline{47.5} & \textbf{48.4} & 18.9 & 57.9 & 1420.7 \\
Figure    & \underline{66.2} & 45.9 & \underline{46.5} & \underline{43.2} & \underline{58.5} & 1482.9 \\
Arxiv     & 65.1 & 46.0 & 45.9 & \textbf{44.1} & \textbf{90.4} & \underline{1490.2} \\
\bottomrule
\end{tabular}
}
\vskip -0.1in
\label{tab:task_order_performance}
\end{wraptable}
 such as FigureQA, we observe backward transfer phenomena,
where learning new tasks enhances the performance of old tasks.
As shown in Figure \ref{fig:mme}, MME scores improve across most tasks in LLaVA-665k. However, during continual fine-tuning, general performance improvements are only observed in Super and Arixv. This may result from differences between these task datasets, as the evaluation does not fully cover task-specific capabilities.

\medskip
\textbf{Comparison of computational overhead.}
Figure \ref{fig:flops} compares the computational cost and training time of these methods on COCO. Continual learning methods introduce additional overhead, requiring longer training time. Our method maintains a comparable cost to previous approaches.

\section{Conclusion}
\label{sec:limit}
To tackle the challenges of task balancing and expansion costs in multitask visual instruction tuning, we present LLaVA-c, a simple but effective continual learning framework.
Our method introduces two key modifications in LLaVA: Spectral-Aware Consolidation (SAC) and Unsupervised Inquiry Regularization (UIR), which address base model degradation and maintain strong plasticity for acquiring new knowledge. 
Experimental results demonstrate that LLaVA-c matches or surpasses multitask training in continual pretraining and maintains general knowledge during task-specific continual finetuning.
We hope this work inspires the application of continual learning in multimodal large language models.

\textbf{Limitations.} In this work, we focus on LLaVA as a representative MLLM due to its implementation efficiency, without exploring other methods. Additionally, our approach has the following limitations due to computational constraints: we use a 7B-scale MLLM without investigating larger models and adopt LoRA instead of full fine-tuning. We ensure that the LoRA configuration aligns with LLaVA-1.5 and merge LoRA parameters after each task to maintain structural consistency with full fine-tuning. Moreover, in our experiments,
most tasks contained around 50K samples, limiting scalability to larger datasets. In the future, we plan to extend this approach to more MLLMs and explore the scale law in continual learning.

\newpage
 {\small  
 \bibliographystyle{unsrt}
 \bibliography{refer}

\begin{thebibliography}{10}

\bibitem{liu2024improved}
Haotian Liu, Chunyuan Li, Yuheng Li, and Yong~Jae Lee.
\newblock Improved baselines with visual instruction tuning.
\newblock In {\em Proceedings of the IEEE/CVF Conference on Computer Vision and Pattern Recognition}, pages 26296--26306, 2024.

\bibitem{brown2020language}
Tom~B Brown, Benjamin Mann, Nick Ryder, Melanie Subbiah, Jared Kaplan, Prafulla Dhariwal, Arvind Neelakantan, Pranav Shyam, Girish Sastry, Amanda Askell, et~al.
\newblock Language models are few-shot learners.
\newblock {\em Advances in neural information processing systems}, 33:1877--1901, 2020.

\bibitem{touvron2023llama}
Hugo Touvron, Louis Martin, Kevin Stone, Peter Albert, Amjad Almahairi, Yasmine Babaei, Nikolay Bashlykov, Soumya Batra, Prajjwal Bhargava, Shruti Bhosale, et~al.
\newblock Llama 2: Open foundation and fine-tuned chat models.
\newblock {\em arXiv preprint arXiv:2307.09288}, 2023.

\bibitem{llava}
Haotian Liu, Chunyuan Li, Qingyang Wu, and Yong~Jae Lee.
\newblock Visual instruction tuning.
\newblock In {\em NeurIPS}, 2023.

\bibitem{radford2021learning}
Alec Radford, Jong~Wook Kim, Chris Hallacy, Aditya Ramesh, Gabriel Goh, Sandhini Agarwal, Girish Sastry, Amanda Askell, Pamela Mishkin, Jack Clark, et~al.
\newblock Learning transferable visual models from natural language supervision.
\newblock {\em arXiv preprint arXiv:2103.00020}, 2021.

\bibitem{kirkpatrick2017overcoming}
James Kirkpatrick, Razvan Pascanu, Neil Rabinowitz, Joel Veness, Guillaume Desjardins, Andrei~A Rusu, Kieran Milan, John Quan, Tiago Ramalho, Agnieszka Grabska-Barwinska, et~al.
\newblock Overcoming catastrophic forgetting in neural networks.
\newblock {\em Proceedings of the National Academy of Sciences}, 114(13):3521--3526, 2017.

\bibitem{chen2024coin}
Cheng Chen, Junchen Zhu, Xu~Luo, Heng~Tao Shen, Jingkuan Song, and Lianli Gao.
\newblock Co{IN}: A benchmark of continual instruction tuning for multimodel large language models.
\newblock In {\em The Thirty-eight Conference on Neural Information Processing Systems Datasets and Benchmarks Track}, 2024.

\bibitem{wang2023orthogonal}
Xiao Wang, Tianze Chen, Qiming Ge, Han Xia, Rong Bao, Rui Zheng, Qi~Zhang, Tao Gui, and Xuan-Jing Huang.
\newblock Orthogonal subspace learning for language model continual learning.
\newblock In {\em Findings of the Association for Computational Linguistics: EMNLP 2023}, pages 10658--10671, 2023.

\bibitem{zeng2024modalprompt}
Fanhu Zeng, Fei Zhu, Haiyang Guo, Xu-Yao Zhang, and Cheng-Lin Liu.
\newblock Modalprompt: Dual-modality guided prompt for continual learning of large multimodal models.
\newblock {\em arXiv preprint arXiv:2410.05849}, 2024.

\bibitem{chowdhery2022palm}
Aakanksha Chowdhery, Sharan Narang, Jacob Devlin, Maarten Bosma, Gaurav Mishra, Adam Roberts, Paul Barham, Hyung~Won Chung, Charles Sutton, Sebastian Gehrmann, et~al.
\newblock Palm: Scaling language modeling with pathways.
\newblock {\em arXiv preprint arXiv:2204.02311}, 2022.

\bibitem{li2023blip}
Junnan Li, Dongxu Li, Silvio Savarese, and Steven Hoi.
\newblock Blip-2: Bootstrapping language-image pre-training with frozen image encoders and large language models.
\newblock {\em arXiv preprint arXiv:2301.12597}, 2023.

\bibitem{zhu2023minigpt4}
Deyao Zhu, Jun Chen, Xiaoqian Shen, Xiangyang Wang, and Ming Ding.
\newblock Minigpt-4: Enhancing vision-language understanding with advanced large language models.
\newblock {\em arXiv preprint arXiv:2304.10592}, 2023.

\bibitem{dai2023instructblip}
Wenliang Dai, Junnan Li, Dongxu Li, Anthony Meng~Huat Tiong, Junqi Zhao, Weisheng Wang, Boyang Li, Pascale Fung, and Steven Hoi.
\newblock Instructblip: Towards general-purpose vision-language models with instruction tuning.
\newblock {\em arXiv preprint arXiv:2305.06500}, 2023.

\bibitem{alayrac2022flamingo}
Jean-Baptiste Alayrac, Jeff Donahue, Pauline Luc, Antoine Miech, Iain Barr, Yana Hasson, Karel Lenc, Arthur Mensch, Katie Millican, Malcolm Reynolds, et~al.
\newblock Flamingo: a visual language model for few-shot learning.
\newblock {\em arXiv preprint arXiv:2204.14198}, 2022.

\bibitem{sun2023emu}
Chen Sun, Cordelia Schmid, Yuxiong He, Hugo Touvron, Armand Joulin, and Antoine Miech.
\newblock Emu: Efficient multimodal pretraining with unified masked prediction.
\newblock {\em arXiv preprint arXiv:2305.04522}, 2023.

\bibitem{li2017learning}
Ang Li, Allan Jabri, Armand Joulin, and Laurens Van Der~Maaten.
\newblock Learning visual n-grams from web data.
\newblock In {\em ICCV}, 2017.

\bibitem{Douillard2020SmallTaskIL}
Arthur Douillard, Matthieu Cord, Charles Ollion, et~al.
\newblock Podnet: Pooled outputs distillation for small-tasks incremental learning.
\newblock In {\em Proceedings of the European Conference on Computer Vision}, pages 86--102, 2020.

\bibitem{wang2022learning}
Zifeng Wang, Zizhao Zhang, Chen-Yu Lee, Han Zhang, Ruoxi Sun, Xiaoqi Ren, Guolong Su, Vincent Perot, Jennifer Dy, and Tomas Pfister.
\newblock Learning to prompt for continual learning.
\newblock In {\em Proceedings of the IEEE/CVF conference on computer vision and pattern recognition}, pages 139--149, 2022.

\bibitem{smith2023coda}
Alexander Smith, Yifan Liu, Hanzhang Peng, Dejing Dou, et~al.
\newblock Codaprompt: Co-designing prompt tuning and architecture for continual learning.
\newblock In {\em International Conference on Learning Representations (ICLR)}, 2023.

\bibitem{Rebuffi2017iCaRLIC}
Sylvestre-Alvise Rebuffi, A.~Kolesnikov, Georg Sperl, and Christoph~H. Lampert.
\newblock icarl: Incremental classifier and representation learning.
\newblock In {\em Proceedings of the IEEE/CVF Conference on Computer Vision and Pattern Recognition}, pages 5533--5542, 2017.

\bibitem{Wu2019LargeSI}
Y.~Wu, Yan-Jia Chen, Lijuan Wang, et~al.
\newblock Large scale incremental learning.
\newblock In {\em Proceedings of the IEEE/CVF Conference on Computer Vision and Pattern Recognition}, pages 374--382, 2019.

\bibitem{gao2022r}
Qiankun Gao, Chen Zhao, Bernard Ghanem, and Jian Zhang.
\newblock R-dfcil: Relation-guided representation learning for data-free class incremental learning.
\newblock In {\em Proceedings of the European Conference on Computer Vision}, pages 423--439. Springer, 2022.

\bibitem{yan2021dynamically}
Shipeng Yan, Jiangwei Xie, and Xuming He.
\newblock Der: Dynamically expandable representation for class incremental learning.
\newblock In {\em Proceedings of the IEEE/CVF Conference on Computer Vision and Pattern Recognition}, pages 3014--3023, 2021.

\bibitem{wang2022foster}
Fu-Yun Wang, Da-Wei Zhou, Han-Jia Ye, and De-Chuan Zhan.
\newblock Foster: Feature boosting and compression for class-incremental learning.
\newblock In {\em Proceedings of the European Conference on Computer Vision}, 2022.

\bibitem{hu2023dense}
Zhiyuan Hu, Yunsheng Li, Jiancheng Lyu, Dashan Gao, and Nuno Vasconcelos.
\newblock Dense network expansion for class incremental learning.
\newblock In {\em Proceedings of the IEEE/CVF Conference on Computer Vision and Pattern Recognition}, pages 11858--11867, 2023.

\bibitem{li2023pope}
Yifan Li, Yifan Du, Kun Zhou, Jinpeng Wang, Wayne~Xin Zhao, and Ji-Rong Wen.
\newblock Evaluating object hallucination in large vision-language models.
\newblock {\em arXiv preprint arXiv:2305.10355}, 2023.

\bibitem{fu2023mme}
Chaoyou Fu, Peixian Chen, Yunhang Shen, Yulei Qin, Mengdan Zhang, Xu~Lin, Zhenyu Qiu, Wei Lin, Jinrui Yang, Xiawu Zheng, et~al.
\newblock Mme: A comprehensive evaluation benchmark for multimodal large language models.
\newblock {\em arXiv preprint arXiv:2306.13394}, 2023.

\bibitem{liu2023mmbench}
Yuan Liu, Haodong Duan, Yuanhan Zhang, Bo~Li, Songyang Zhang, Wangbo Zhao, Yike Yuan, Jiaqi Wang, Conghui He, Ziwei Liu, et~al.
\newblock Mmbench: Is your multi-modal model an all-around player?
\newblock {\em arXiv preprint arXiv:2307.06281}, 2023.

\bibitem{li2023seed}
Bohao Li, Rui Wang, Guangzhi Wang, Yuying Ge, Yixiao Ge, and Ying Shan.
\newblock Seed-bench: Benchmarking multimodal llms with generative comprehension.
\newblock {\em arXiv preprint arXiv:2307.16125}, 2023.

\bibitem{yu2023mmvet}
Weihao Yu, Zhengyuan Yang, Linjie Li, Jianfeng Wang, Kevin Lin, Zicheng Liu, Xinchao Wang, and Lijuan Wang.
\newblock Mm-vet: Evaluating large multimodal models for integrated capabilities.
\newblock {\em arXiv preprint arXiv:2308.02490}, 2023.

\bibitem{yue2024mmmu}
Xiang Yue, Yuansheng Ni, Kai Zhang, Tianyu Zheng, Ruoqi Liu, Ge~Zhang, Samuel Stevens, Dongfu Jiang, Weiming Ren, Yuxuan Sun, et~al.
\newblock Mmmu: A massive multi-discipline multimodal understanding and reasoning benchmark for expert agi.
\newblock In {\em Proceedings of the IEEE/CVF Conference on Computer Vision and Pattern Recognition}, pages 9556--9567, 2024.

\bibitem{li2023super}
Zhuowan Li, Xingrui Wang, Elias Stengel-Eskin, Adam Kortylewski, Wufei Ma, Benjamin Van~Durme, and Alan~L Yuille.
\newblock Super-clevr: A virtual benchmark to diagnose domain robustness in visual reasoning.
\newblock In {\em Proceedings of the IEEE/CVF Conference on Computer Vision and Pattern Recognition}, pages 14963--14973, 2023.

\bibitem{lu2021iconqa}
Pan Lu, Liang Qiu, Jiaqi Chen, Tony Xia, Yizhou Zhao, Wei Zhang, Zhou Yu, Xiaodan Liang, and Song-Chun Zhu.
\newblock Iconqa: A new benchmark for abstract diagram understanding and visual language reasoning.
\newblock In {\em The 35th Conference on Neural Information Processing Systems (NeurIPS 2021) Track on Datasets and Benchmarks}, 2021.

\bibitem{dahlgren2022clevr}
Adam Dahlgren~Lindstr{\"o}m and Savitha Sam~Abraham.
\newblock Clevr-math: A dataset for compositional language, visual and mathematical reasoning.
\newblock In {\em International Joint Conference on Learning and Reasoning, 16th International Workshop on Neural-Symbolic Learning and Reasoning (NeSy 2022), Windsor, UK, September 28-30, 2022}, volume 3212, pages 155--170. Technical University of Aachen, 2022.

\bibitem{li-etal-2024-multimodal-arxiv}
Lei Li, Yuqi Wang, Runxin Xu, Peiyi Wang, Xiachong Feng, Lingpeng Kong, and Qi~Liu.
\newblock Multimodal {A}r{X}iv: A dataset for improving scientific comprehension of large vision-language models.
\newblock In Lun-Wei Ku, Andre Martins, and Vivek Srikumar, editors, {\em Proceedings of the 62nd Annual Meeting of the Association for Computational Linguistics (Volume 1: Long Papers)}, pages 14369--14387, Bangkok, Thailand, August 2024. Association for Computational Linguistics.

\bibitem{kahou2017figureqa}
Samira~Ebrahimi Kahou, Vincent Michalski, Adam Atkinson, {\'A}kos K{\'a}d{\'a}r, Adam Trischler, and Yoshua Bengio.
\newblock Figureqa: An annotated figure dataset for visual reasoning.
\newblock {\em arXiv preprint arXiv:1710.07300}, 2017.

\bibitem{zheng2023judging}
Lianmin Zheng, Wei-Lin Chiang, Ying Sheng, Siyuan Zhuang, Zhanghao Wu, Yonghao Zhuang, Zi~Lin, Zhuohan Li, Dacheng Li, Eric Xing, et~al.
\newblock Judging llm-as-a-judge with mt-bench and chatbot arena.
\newblock {\em Advances in Neural Information Processing Systems}, 36:46595--46623, 2023.

\end{thebibliography}
 }
 







\newpage
\appendix
\section{Appendix}

\subsection{Additional Implementation  Details}
\label{other-details}
LLaVA, LLaVA-1.5, and LLaVA-1.5c share the same set of hyperparameters. The only difference is in the vision-language alignment stage, where LLaVA-1.5 reduces the learning rate by half compared to LLaVA. The authors claim that this adjustment is due to using an MLP projection layer instead of the original linear projection layer design~\cite{liu2024improved}.
Our method does not introduce any modifications to the vision-language alignment stage. Table~\ref{tab:llava15_hparams} presents the training hyperparameters for both vision-language alignment and the second-stage visual instruction tuning. Our method utilizes LoRA fine-tuning, maintaining the same hyperparameter settings as LLaVA-1.5.

\begin{table}[h]
    \centering
    \caption{Training hyperparameters of LLaVA-c.}
    \vskip 0.05in
    \label{tab:llava15_hparams}
    \small
\centering
\renewcommand\tabcolsep{5pt}
\renewcommand{\arraystretch}{1.2}
    \resizebox{0.4\linewidth}{!}{ 
    \begin{tabular}{l|cc}
    \toprule
    \textbf{Hyperparameter} & \textbf{Alignment} & \textbf{Finetune} \\
    \midrule
    batch size & 256 & 128 \\
    learning rate & 1e-3 & 2e-5 \\
    lr schedule & \multicolumn{2}{c}{cosine decay} \\
    lr warmup ratio & \multicolumn{2}{c}{0.03} \\
    weight decay & \multicolumn{2}{c}{0} \\
    epoch & \multicolumn{2}{c}{1} \\
    optimizer & \multicolumn{2}{c}{AdamW} \\
    DeepSpeed stage &zero2 &zero3 \\
    LoRA $R$ &\multicolumn{2}{c}{128} \\
    LoRA $\alpha$ &\multicolumn{2}{c}{256} \\
    \bottomrule
    \end{tabular}
    }
\end{table}

\subsection{Dataset Details}
\label{app-details}
This subsection provides a detailed overview of the datasets used for continual pretraining and fine-tuning. For continual pretraining, we categorize the data into six tasks based on different data sources within the LLava-665k dataset. TQA is a purely text-based question-answering dataset, and we extract questions to serve as an unsupervised inquiry dataset.
Numerous task-specific datasets can be used for continual fine-tuning experiments. We select five datasets where LLaVA-1.5 exhibits poor zero-shot performance. We randomly sample subsets from each of these datasets and divide them into training and test sets. To assess model efficiency, we set the test set size to 2,000 samples, as we found that evaluation accuracy on 10,000 samples is nearly identical to that on 2,000 samples. Table \ref{tab:datasets} presents the number of samples in these datasets and the task-specific training instructions.

\begin{table}[h]
    \caption{Details of datasets used in continual pretraining and continual fine-tuning.}
        \label{tab:datasets}
    \vskip 0.05in
    \small
    \centering
    \renewcommand\tabcolsep{5pt}
    \renewcommand{\arraystretch}{1.2}
    \resizebox{0.95\linewidth}{!}{ 
    \begin{tabular}{lcc|c}
    \toprule
    \textbf{Datasets} & \textbf{Train} & \textbf{Test} & \textbf{Task-specific Instruction} \\
    \midrule
    \multicolumn{3}{l|}{\textit{Continual Pretraining:}} \\
    TQA & 40688 & - & - \\
    COCO & 364100 & - & - \\
    OCRQA & 80000 & - & Answer the question using a single word or phrase. \\
    TextVQA & 21953 & - & Provide a one-sentence caption for the image. \\
    GQA & 72140 & - & Answer the question using a single word or phrase. \\
    VG & 86417 & - & Provide the bounding box coordinate of the region. \\
    \midrule
    \multicolumn{3}{l|}{\textit{Continual Fine-tuning:}} \\
    IconQA & 29859 & 2000 & Answer with the option's letter from the given choices directly.  \\
    Super & 20000 & 2000 & Answer the question using a single word or phrase. \\
    Clevr-Math & 40000 & 2000 & Answer the question using a single word or phrase. \\
    FigureQA & 13597 & 2000 & Please answer the question and provide the correct option letter at the end. \\
    ArxivQA & 40000 & 2000 & Answer with the option's letter from the given choices directly. \\
    \bottomrule
    \end{tabular}
    }

\end{table}

\subsection{Examples of Tasks}
As shown in Table \ref{tab:demo}, we present examples from various tasks in Llava 665k, which share similar image domains, except for OCR. These tasks cover diverse VQA types, including visual grounding, caption generation, and more.
Our approach learns these tasks sequentially, resulting in a strong overall MLLM performance. Since the image domains in these tasks differ significantly from those seen during continual fine-tuning, zero-shot performance is suboptimal. Nevertheless, our method consistently improves performance across both natural images and specialized domains, such as IconQA and ArxivQA, demonstrating its effectiveness in continual learning.

\begin{table}[t]
\centering
\caption{Task Data with Images Across \textit{COCO}, \textit{OCR}, \textit{Text
QA}, \textit{GQA}, and \textit{VG}}
\label{tab:demo}
\vskip 0.05in
\renewcommand\tabcolsep{5pt}
\renewcommand{\arraystretch}{1.2}
\resizebox{0.95\linewidth}{!}{ 
\begin{tabular}{m{8cm}| m{5cm}} 
\toprule
\adjustbox{center}{\includegraphics[width=0.5\linewidth]{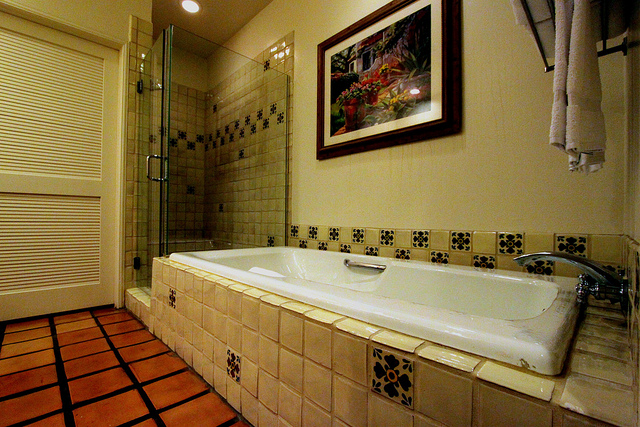}} &  
\textit{\color{blue}Question:} What objects are featured in the bathroom shown in the image? \newline \newline   
\textit{\color{blue}Answer:} In the bathroom shown in the image, there is a bathtub, a separate shower stall, a towel, and a framed picture. \\
\midrule
\adjustbox{center}{\includegraphics[width=0.26\linewidth]{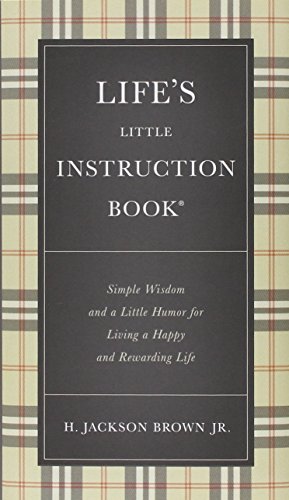}} &  
\textit{\color{blue}Question:} Who is the author of this book? Answer the question using a single word or phrase. \newline 
\textit{\color{blue}Answer:} H. Jackson Brown \\
\midrule
\adjustbox{center}{\includegraphics[width=0.5\linewidth]{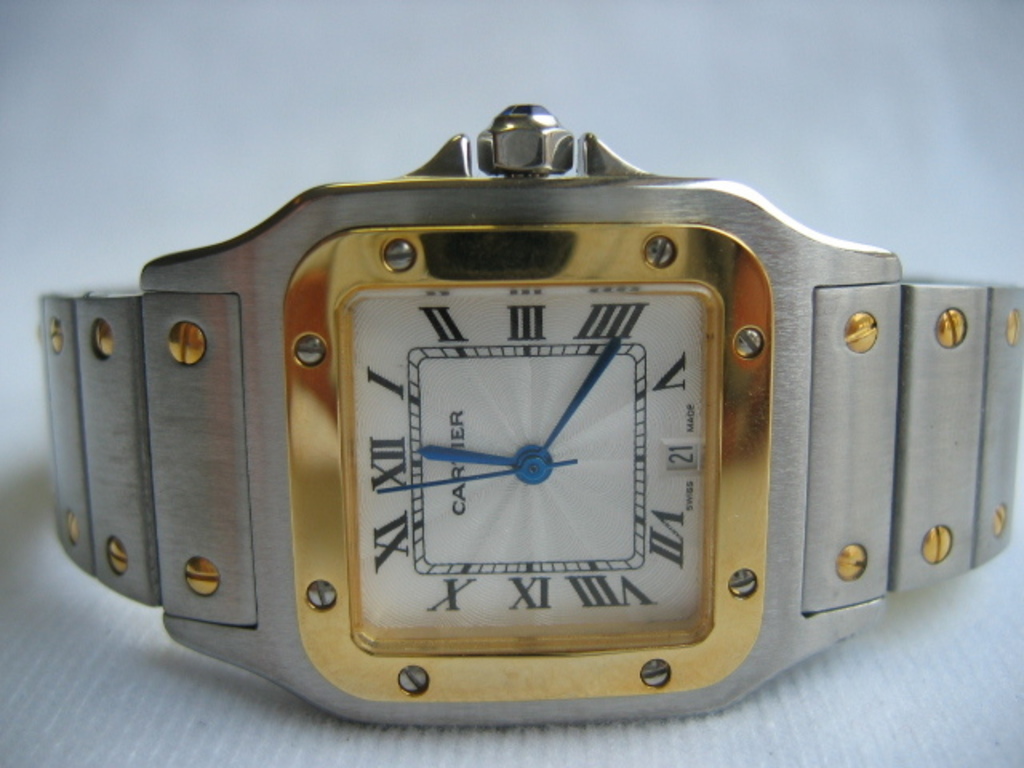}} &  
\textit{\color{blue}Question:} Provide a one-sentence caption for the provided image. Reference OCR token: III, SIITIIILD, X, XI, IA \newline \newline   
\textit{\color{blue}Answer:} A gold cartier watch is on its side showing the time is approaching 12.22. \\
\midrule
\adjustbox{center}{\includegraphics[width=0.5\linewidth]{demo/2315372.jpg}} &  
\textit{\color{blue}Question:} Is the bag to the left or to the right of the girl that is wearing pants? Answer the question using a single word or phrase. \newline \newline   
\textit{\color{blue}Answer:} Right. \\
\midrule
\adjustbox{center}{\includegraphics[width=0.5\linewidth]{demo/8.jpg}} &  
\textit{\color{blue}Question:} Please provide the bounding box coordinate of the region this sentence describes: dish is ontop of cpu. \newline \newline   
\textit{\color{blue}Answer:} [0.22, 0.42, 0.26, 0.48]. \\
\bottomrule
\end{tabular}
}
\end{table}

\subsection{Training Process of LLaVA-c}
\label{app-code}
The complete continual learning process of LLaVA-c is detailed in Algorithm~\ref{alg:llava-c}, which consists of two key phases in each task iteration:
(1) Learning the new task with Unsupervised Inquiry Regularization (UIR).
(2) Integrating new and old knowledge through Spectral-Aware Consolidation.
Our approach introduces no additional hyperparameters during visual instruction tuning compared to vanilla LLaVA and LLaVA-1.5. The UIR loss is directly added to the original LLaVA loss, making it simple to implement.
Furthermore, Spectral-Aware Consolidation iterates over parameter updates and adaptively scales eigenvalues to balance knowledge retention and adaptation. Our method will be fully open-sourced.

\begin{algorithm}[t]
  \caption{Continual Learning Process of LLaVA-c}
  \label{alg:llava-c}
  \begin{algorithmic}[1]
    \REQUIRE Initial model $\bm{\theta}_0$, task sequence $\{\mathcal{D}_1, \mathcal{D}_2, \dots, \mathcal{D}_T\}$
    \ENSURE  Consolidated model $\widetilde{\bm{\theta}}_T$
    \FOR{$t = 1$ to $T$}
      \STATE \textbf{\color{gray}$\triangleright$ Task Training:}
      \STATE Calculate 
        $
          \mathcal{L}^{t}_{\text{llava}} = \mathbb{E}_{\mathcal{D}_t}[\ell(f_{\bm{\theta}}(\bm{x}), \bm{y})],\quad
          \mathcal{L}^{t}_{\text{uir}} = \mathbb{E}_{\bm{x} \sim \mathcal{D}_{\text{inquiry}}}\bigl\|h[f_{\bm{\theta}_{t}}(\bm{x})]-h[f_{\bm{\theta}_{t-1}}(\bm{x})]\bigr\|_2
        $
      \STATE Optimize 
        $
          \bm{\theta}_t = \arg\min_{\bm{\theta}} \bigl(\mathcal{L}^{t}_{\text{llava}} + \mathcal{L}^{t}_{\text{uir}}\bigr)$
      \newline
      \STATE \textbf{\color{gray}$\triangleright$ Spectral-Aware Consolidation:}
      \STATE Compute update $\Delta\bm{\theta} = \bm{\theta}_t - \bm{\theta}_{t-1}$
      \STATE Decompose $\Delta\bm{\theta} = \mathbf{U}\,\mathbf{\Sigma}\,\mathbf{V}^\top$
      \FOR{$i = 1$ to $r$}
        \STATE window size 
          $k = \arg\min_{k}\Bigl(\tfrac{1}{k}\sum_{j=1}^{\min(k+1,r)}\sigma_j - \alpha\sigma_1\Bigr)^2$
        \STATE Compute 
          $ 
            \widetilde\sigma_i = \frac{1}{k}\sum_{j=i}^{\min(i+k,r)}\sigma_j$
      \ENDFOR
      \STATE Reconstruct 
        $
          \Delta\widetilde{\bm{\theta}} = \mathbf{U}\,\mathrm{diag}(\widetilde\sigma_1,\ldots,\widetilde\sigma_r)\,\mathbf{V}^\top$
      \STATE Update 
        $\widetilde{\bm{\theta}}_t = \bm{\theta}_{t-1} + \Delta\widetilde{\bm{\theta}}$
    \ENDFOR
  \end{algorithmic}
\end{algorithm}

\subsection{More experimental results}

\noindent\textbf{Comparison with ModelMix.} Figure \ref{fig:modelmix} presents a comprehensive comparison with ModelMix across multiple datasets, demonstrating that our method significantly outperforms ModelMix. Tables \ref{app:modelmix} summarize the results under continual pretraining settings, which are consistent with the improvements shown in Figure \ref{fig:modelmix}.
\begin{table}[h]
\vskip -0.1in
\centering
\setlength\tabcolsep{6pt}
\renewcommand{\arraystretch}{1.05}
\caption{Comparison with ModelMix under continual pretraining}
\vskip 0.05in
\label{app:modelmix}
\resizebox{0.85\linewidth}{!}{
\begin{tabular}{l c c c c c c c c c}
\toprule
\multirow{2}{*}{Model} & \multirow{2}{*}{MME} & \multicolumn{2}{c}{MMBench} & \multirow{2}{*}{SEED-Bench} & \multirow{2}{*}{LLaVA-Wild} & \multirow{2}{*}{MM-Vet} & \multicolumn{3}{c}{POPE} \\
 & & en & cn & & & & rand & pop & adv \\
\midrule
ModelMix   & 1358.2 & 50.1 & 47.6 & 37.2 & 29.7 & 26.5 & 84.2 & 82.9 & 82.0 \\
\rowcolor{Gray}
Our & \textbf{1502.4} & \textbf{65.2} & \textbf{58.8} & \textbf{61.0} & \textbf{66.8} & \textbf{31.2} & \textbf{87.7} & \textbf{86.2} & \textbf{85.1} \\
\bottomrule
\end{tabular}
}
\vskip -0.1in
\end{table}

\medskip
\noindent\textbf{Comparison with full data-replay.}
In our ablation study (Figure \ref{fig:data-replay}), we add unsupervised queries to the replay buffer and assign them synthetic target labels to meet SFT requirements. We also evaluate full data replay by jointly training the base dataset (LLaVA-665k) with all new task data (IconQA, Super, Math, Figure, Arxiv); the results are reported in Table \ref{tab:training-time-comparison}.
\begin{table}[h]
\vskip -0.1in
\centering
\setlength\tabcolsep{6pt}
\renewcommand{\arraystretch}{1.1}
\caption{Comparison with full data-replay training}
\vskip 0.05in
\label{tab:training-time-comparison}
\resizebox{0.75\linewidth}{!}{
\begin{tabular}{l c c c c c c c}
\toprule
\multirow{2}{*}{Method} 
  & \multicolumn{6}{c}{Benchmarks} 
  & \multirow{2}{*}{Time (h)} \\
\cmidrule(lr){2-7}
  & IconQA & Super & Math & Figure & Arxiv & LLaVA-bench & \\
\midrule
Data-replay (all) 
  & \textbf{69.4}   & 44.5  & 43.2 & 42.8   & 86.9  & \textbf{64.6}        & 34.7         \\

\rowcolor{Gray}
Our       
  & 65.1 & \textbf{46.0} & \textbf{45.9} & \textbf{44.1} & \textbf{90.4} & 62.4 & \textbf{4.8} \\
\bottomrule
\end{tabular}
}
\vskip -0.1in
\end{table}

Compared to data-replay retraining, our method \textbf{(1)} requires substantially less time while matching full data-replay performance on downstream tasks, and \textbf{(2)} is especially practical since the original training data for most open-source MLLMs is unavailable.

\medskip
\noindent\textbf{Experiment on more models.}
Building on the LLaVA-1.5 architecture, we conducted continual fine-tuning on several LLM backbones. The results in Table \ref{app:backbone} confirm that our approach yields significant and consistent improvements across these models.

\begin{table}[h]
\vskip -0.1in
\centering
\setlength\tabcolsep{7pt}
\renewcommand{\arraystretch}{1.2}
\caption{{Comparison of different models under continual fine-tuning}}  
\vskip 0.05in
\label{app:backbone}
\resizebox{0.66\linewidth}{!}{%
\begin{tabular}{@{}llcccccc@{}}
\toprule
Backbone    & Method  & IconQA & Super & Math  & Figure & Arxiv & LLaVA-bench \\
\midrule
\multirow{2}{*}{LLaMA2-7B}   & Vanilla  &  9.7   & 33.1  & 25.2  & 35.8   & 90.3  & 22.3        \\
  & {\cellcolor{Gray}Our}     
            & {\cellcolor{Gray}\textbf{63.2}} 
            & {\cellcolor{Gray}\textbf{44.9}} 
            & {\cellcolor{Gray}\textbf{43.5}} 
            & {\cellcolor{Gray}\textbf{40.7}} 
            & {\cellcolor{Gray}\textbf{91.0}} 
            & {\cellcolor{Gray}\textbf{56.4}} \\
\midrule
\multirow{2}{*}{Vicuna-7B}   & Vanilla  & 11.3   & 35.1  & 27.5  & 38.5   & 92.6  & 25.0        \\
   & {\cellcolor{Gray}Our}     
            & {\cellcolor{Gray}\textbf{65.1}} 
            & {\cellcolor{Gray}\textbf{46.0}} 
            & {\cellcolor{Gray}\textbf{45.9}} 
            & {\cellcolor{Gray}\textbf{44.1}} 
            & {\cellcolor{Gray}\textbf{90.4}} 
            & {\cellcolor{Gray}\textbf{62.4}} \\
\midrule
\multirow{2}{*}{Mistral-7B}  & Vanilla  & 20.9   & 37.8  & 33.7  & 34.6   & 95.0  & 42.7        \\
  & {\cellcolor{Gray}Our}     
            & {\cellcolor{Gray}\textbf{70.1}} 
            & {\cellcolor{Gray}\textbf{55.3}} 
            & {\cellcolor{Gray}\textbf{52.6}} 
            & {\cellcolor{Gray}\textbf{52.7}} 
            & {\cellcolor{Gray}\textbf{91.4}} 
            & {\cellcolor{Gray}\textbf{76.5}} \\
\bottomrule
\end{tabular}%
}
\vskip -0.1in
\end{table}

\medskip
\noindent\textbf{Ablation study on $\alpha$.}
In Figure \ref{fig:modelmix}, we plot the $\alpha$-performance curves under continual fine-tuning for the first task (IconQA) and the final task (ArxivQA). Below, we show the corresponding curves for the intermediate tasks (Super, Math, and Figure):

\begin{figure*}[h]
\centering
\includegraphics[width=1.0\linewidth]{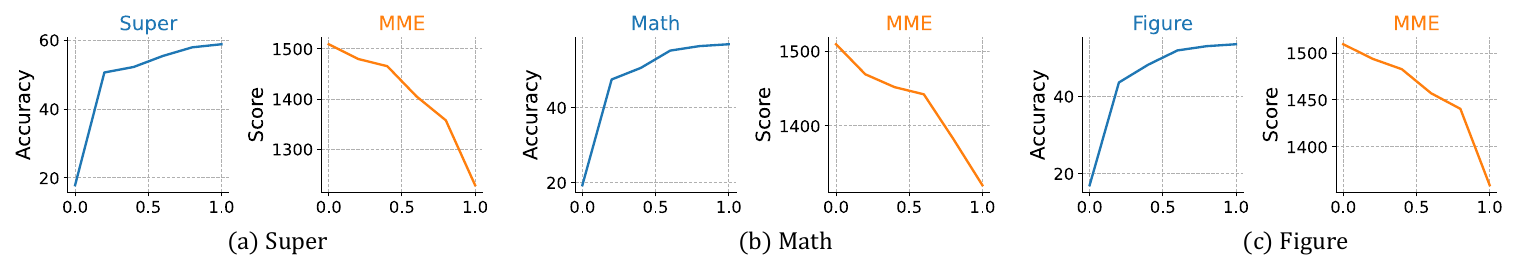}
\vskip -0.05in
\caption{$\alpha$-performance curves for intermediate tasks under continual fine-tuning.}

\label{app-fig}
\end{figure*}

Compared to multi-task data mixing, SAC-coefficient tuning uses a single parameter, operates on CPU with minimal overhead, and can explore the entire range in minutes. Data mixing, however, adds $N-1$ hyperparameters for $N$ tasks—resulting in an exponentially growing search space—and forces full retraining (data reintegration, hyperparameter optimization, and model retraining) whenever a new task arrives.


\end{document}